\documentclass[journal]{IEEEtran}
\IEEEoverridecommandlockouts
\usepackage{cite}
\usepackage{amsmath,amssymb,amsfonts}
\usepackage{algorithmic}
\usepackage{graphicx}
\usepackage{textcomp}
\usepackage{xcolor}
\usepackage{gensymb}
\usepackage{wrapfig}
\usepackage{subcaption}  
\usepackage{makecell}
\usepackage[numbers]{natbib}
\usepackage{multirow}
\def\BibTeX{{\rm B\kern-.05em{\sc i\kern-.025em b}\kern-.08em
    T\kern-.1667em\lower.7ex\hbox{E}\kern-.125emX}}

\begin{document}

\title{CorrFL: Correlation-based Neural Network Architecture for Unavailability Concerns in a Heterogeneous IoT Environment}


\author{
    \IEEEauthorblockN{Ibrahim Shaer\IEEEauthorrefmark{1}, Abdallah Shami\IEEEauthorrefmark{1}} \\
    \IEEEauthorblockA{\IEEEauthorrefmark{1}Department of Electrical and Computer Engineering \\
    Western University \\
    London ON, Canada 
    \\\{ishaer, abdallah.shami\}@uwo.ca}
}

\markboth{Manuscript Published in IEEE Transactions on Network and Service Management}%
{Manuscript Published in IEEE Transactions on Network and Service Management}
\maketitle
\begin{abstract}
The Federated Learning (FL) paradigm faces several challenges that limit its application in real-world environments. These challenges include the local models' architecture heterogeneity and the unavailability of distributed Internet of Things (IoT) nodes due to connectivity problems. These factors posit the question of ``how can the available models fill the training gap of the unavailable models?". This question is referred to as the ``Oblique Federated Learning" problem. This problem is encountered in the studied environment that includes distributed IoT nodes responsible for predicting CO\textsubscript{2} concentrations. This paper proposes the Correlation-based FL (CorrFL) approach influenced by the representational learning field to address this problem.   CorrFL projects the various model weights to a common latent space to address the model heterogeneity. Its loss function minimizes the reconstruction loss when models are absent and maximizes the correlation between the generated models. The latter factor is critical because of the intersection of the feature spaces of the IoT devices. CorrFL is evaluated on a realistic use case, involving the unavailability of one IoT device and heightened activity levels that reflect occupancy. The generated CorrFL models for the unavailable IoT device from the available ones trained on the new environment are compared against models trained on different use cases, referred to as the benchmark model. The evaluation criteria combine the mean absolute error (MAE) of predictions and the impact of the amount of exchanged data on the prediction performance improvement. Through a comprehensive experimental procedure, the CorrFL model outperformed the benchmark model in every criterion. 
\end{abstract}


\section{Introduction}
%
%
%
%

The evolution of sensing and computing technologies spearheaded the rise of inter-connected devices, known as the Internet of Things (IoT). These devices, which include sensors and edge nodes, are tasked with collecting data from the ambient environment to facilitate the automation systems' decision-making process \cite{rose2015internet}. Since each IoT device captures an aspect of the environment, a centralized server can be employed to gather all IoT device data and create Machine Learning (ML) models to realize a specific task. However, this centralized paradigm is faced with many challenges. These challenges include bandwidth limitations, exposing the environment to a single point of failure, data privacy concerns, and the availability of individual sensors \cite{shaer2020multi, kairouz2021advances}.

Federated Learning (FL) \cite{kairouz2021advances} is a decentralized paradigm that addresses these challenges by collaboratively training an ML model using local agents' models. This paradigm follows a two-stage process. The local agents train their models using their collected data, and their corresponding model weights are dispatched to the central server. After that, the central server aggregates these weights and transmits the updated weights to the local agents. Lastly, each local agent updates its model weights using the shared model's weights. However, this crude implementation faces several hurdles hampering its implementation in real-world environments. Despite the corpus of works discussing and addressing salient issues in the FL environment, such as vertical FL \cite{wahab2021federated}, the stragglers issue \cite{arafeh2022data}, and the heterogeneity in sample space \cite{arafeh2022independent}, the joint consideration of participants' model heterogeneity and the availability of local agents have eluded the research community. In what follows, the explanation of these challenges and how they are encountered in real-world scenarios is detailed. 

\begin{figure}
    \centering
\includegraphics[scale = 0.3]{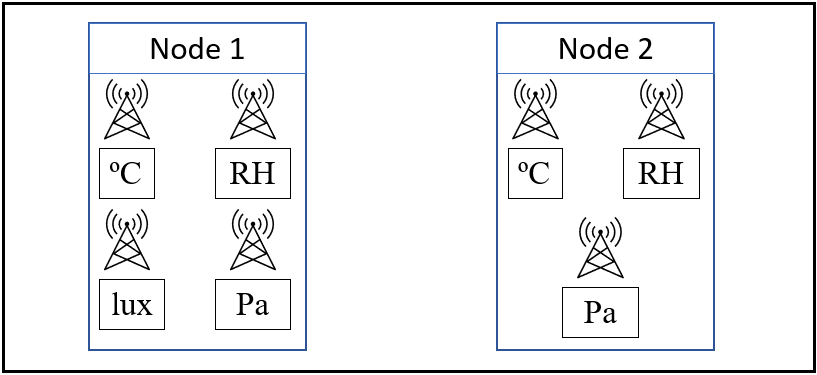}
    \caption{Illustrative Example}
    \label{fig:illustrative_example}
\end{figure}

\textbf{Model Heterogeneity Challenge}: The model heterogeneity refers to the non-uniformity of the feature space fed to the Neural Network (NN) model. The feature space of local agents differs due to two main reasons. First, the feature space of local agents consisting of IoT devices incorporates sensors of different capabilities, potentially capturing different sets of complementary features. This fact is encouraged by the constant enhancement of sensors' battery lifetime, compelling the stakeholders to deploy new devices from various manufacturers each with their own set of standards for IoT devices \cite{callebaut2021art}. An illustrative example of this challenge is the deployment of two IoT devices with different readings depicted in Figure \ref{fig:illustrative_example}. Node 2 captures three readings of temperature (\degree C), humidity (RH), and pressure (Pa), while Node 1 captures four readings of temperature  (\degree C), humidity (RH), light intensity (lux) and pressure (Pa). Second, IoT devices can be equipped with heterogeneous resources, a condition reflected in the generated models in two ways. It limits their ability to capture specific ambient features, creating a disparity in the features fed to the NN model, a factor touched upon in the first reason for model heterogeneity. The second way to address this hurdle is capturing the intended environment space using all the available sensors, albeit catering the feature engineering technique to the resources available on each IoT device. Both of these methods are meant to address the stragglers problem in the FL paradigm research space. As a result of these factors, each IoT device captures different sets of features that share some features. The downstream effect is generating models with heterogeneous architectures. 

\textbf{Availability of Local Agents}: The availability of local agents and their updated model weights is in jeopardy, due to connectivity and energy constraints of IoT devices \cite{lim2020federated}. The effect of participants' dropout is aggravated when no redundancy is incorporated into the FL environment, which means that the local models with their unique set of features would not be updated in downtime scenarios \cite{macedo2014dependability}. This scenario creates a divergence between the set of available and unavailable models.

These combined factors require contingency plans on the aggregation server's end to keep generating models so that each local model, including the unavailable ones, can continue its training process. These factors undermine the application of vanilla FL paradigms that assume the uniformity of model architectures (e.g. all models use the same set of features) and the availability of local agents.


The model heterogeneity challenges have been previously addressed in the literature using knowledge distillation, for example in the works of \cite{lin2020ensemble, chang2019cronus, li2019fedmd}. The knowledge distillation method necessitates the existence of a public dataset sent to each local agent. Each agent is trained using these public datasets, and their classification results are transmitted to the central server to infer the underlying models' architecture. The literature has adopted different methodologies to combine these results to generate a shared global model.  These methods are limited by the need to craft a public dataset and the lack of incorporating unavailability aspects in their formulation.

These factors prompt the research question that we answer in the manuscript, ``How can we leverage the knowledge of a set of models to fill the training gap of the unavailable model when sharing some of its feature space?". The question combining the environment's assumptions and the outlined challenges is coined ``Oblique Federated Learning: Learning from other participants". ``Oblique Federated Learning" assumes that the local agents share common features; however, they do not capture the exact same features. ``Oblique Federated Learning" is halfway between Horizontal FL which assumes the agents' feature space uniformity and Vertical FL which assumes that the agents' feature space is formed of disjoint sets. 

\textbf{Use Cases}: To bridge the gap between the hypothetical scenarios outlined in the challenges and the real-world scenarios, it is imperative to present some use cases that convey these challenges. These use cases target two applications in the field of autonomous vehicles and intrusion detection systems. Radar interference management is a prerequisite for the ubiquitous deployment of radar technology to enable autonomous vehicles to mitigate the dangerous blinding of their radars in dense urban communities \cite{yaqoob2019autonomous}. Since radar technology is essential for the realization of autonomous vehicle applications, the blinding phenomenon reflects the non-uniformity of the feature spaces of vehicles in an FL application targeting scene detection. ML-based solutions for intrusion detection attacks utilize a set of features representing the network traffic to predict the probability of an attack. However, some sets of these features can be unavailable, due to privacy concerns related to each organization or uninformative, due to the masking of some features (i.e., IP addresses might be unavailable). These two examples illustrate the pervasiveness of the ``Oblique Federated Learning” phenomenon. 

In this work, we address the “Oblique Federated Learning” phenomenon in a distributed IoT environment, consisting of sets of sensors that gather different environmental features and suffer from availability constraints. These devices are responsible for predicting CO\textsubscript{2} concentrations over a future time horizon. This dataset was chosen since the model heterogeneity is inherent in its non-uniform feature space, which does not necessitate forging hypothetical scenarios. Moreover, this application's utility is reflected in different domains. CO\textsubscript{2} concentrations can act as proxy estimators of occupancy, aiding the Heating, Ventilation, and Air Conditioning (HVAC) systems in their decision-making process. As a result, a more informed HVAC control is obtained, which improves occupants' comfort \cite{satish2012co2}, curbs the spread of COVID-19 \cite{fadaei2021ventilation}, and reduces energy consumption \cite{manning2007effects}. 

In the studied environment, each IoT device collects some common features with other IoT devices, which means that their corresponding NN models share some neuron combinations. The central aggregation server can leverage this fact to generate updated models for the local agents when they cannot send their updated model weights. Toward that end, this paper proposes a novel NN model, termed as Correlational FL (CorrFL), inspired by the representational learning literature, Correlational Neural Networks in particular \cite{chandar2016correlational}, to produce updated NN weights for unavailable IoT devices from the available ones.

This paper evaluates the proposed CorrFL method in various use-case scenarios of activity levels associated with CO\textsubscript{2} concentration changes. Under minimal or no activity levels, the distributed IoT devices are trained using conventional FL paradigms. One of the local models becomes unavailable when heightened activity levels are encountered. In this case, the CorrFL is employed to generate models for the absent local agents to demonstrate its ability to mitigate their absence. The generated models' performance is compared to those obtained using only the training process, referred to as the benchmark models. 


In summary, the contributions of this paper are as follows:
\begin{itemize}
    
    \item Coin the term ``Oblique Federated Learning" to combine model heterogeneity and availability concerns in an FL environment;
    \item Propose a novel NN architecture that expands the Correlational Neural Networks to a multi-view environment, representing different models, and amends its loss function to tailor to the requirements of the studied environment;
    \item Devise a loss function that incorporates the unavailability of models and maximizes the correlations between the generated models;
    
    \item Propose an evaluation criterion that combines networking concerns and prediction performance referred to as data size exchange per one percentage prediction improvement; and,
    
    \item Introduce the concepts of ``delay," representing the period between the start of local agents' training and the start of central server training and ``Model Dispatch Frequency," describing the number of local model weights sent at once, in the FL environment and highlight their importance on both the network perspective and the accuracy of CO\textsubscript{2} concentration predictions.
\end{itemize}

The rest of the paper is organized as follows. Section II introduces background information pertaining to the proposed approach. Section III details the related work and discusses its limitations in relation to the adopted approach. Section IV explains the components of the devised methodology. Section V outlines the experimental parameters. Section VI explains the use case scenario and its motivation. Section VII details the results. Section VIII concludes the paper. 

\section{Background}

A Correlational Neural Network (CorrNet) \cite{chandar2016correlational} is a staple implementation in the field of representational learning. CorrNet is built on a premise that data, pertaining to a common environment, consists of different views correlated by their connection to the bespoke environment. For example, a movie can be decomposed into a series of images, audio signals, and subtitles. As such, it is expected that a correlation exists between these views, as they represent a movie's modality. Therefore, when one of these modalities is in-existent, other views should construct the missing view. The CorrNet architecture and its loss functions reflect these concerns. 

The different modalities of a single view consist of distinct dimensions. Therefore, the CorrNet projects the views into a common dimensionality.  With their encoder and decoder components, Autoencoders (AE) \cite{baldi2012autoencoders} can provide this function by projecting each view into a common subspace. In the CorrNet implementation, the latent space views are added to obtain a common representation of all views. This common representation is of the same dimension as the latent space. After that, the decoder reconstructs the original views based on a composite loss function.

The main requirement of reconstructing an absent view from other views is incorporated into CorrNet's loss function. The loss function encompasses three main concerns conveyed by L1 loss, L2 loss, and L3 loss. The L1 loss is the reconstruction loss that is similar to AE architectures' loss. The L2 loss is the reconstruction loss of the views when one of them is missing. The number of L2 losses is equivalent to the number of views. Lastly, the L3 loss calculates the correlation between common representations when one of the views is missing and deducts its value from the reconstruction losses. 

\section{Related Work}
This section covers works related to the applied application, global model adaptations to the FL paradigm's shortcomings and NN similarity inference to highlight this work's novelty. 

The early adoption of FL in literature involved a basic implementation such that each client shares the same feature space and different sample space, referred to as the horizontal FL \cite{arafeh2023modularfed}. Horizontal FL has been applied to different use cases in the field of IoT applications, especially in the fields of energy predictions, and smart buildings. For example, the work of Saptura \textit{et al.} \cite{saputra2019energy} investigates the application of FL to predict Electric Vehicle (EV) energy demand. In the context of smart buildings, the FL paradigm is widely adopted to predict energy consumption. Examples of these applications include \cite{wang2022personalized, husnoo2022fedrep, li2022federated, gholizadeh2022federated, savi2021short} that share the same purpose, albeit with some prominent differences in the applied methodology. A common theme in these works is to group local clients to minimize the number of participants or personalize the energy consumption profile. The works by \cite{savi2021short, wang2022personalized} cluster users based on their energy consumption profile whereas Gholizadeh \textit{et al.} \cite{gholizadeh2022federated} groups users using their hyper-parameter optimization results that reveal this similarity. The works \cite{li2022federated, husnoo2022fedrep} plainly predict the energy consumption profiles in different environments. 

Many literary works extend the conventional FL paradigm to adapt to different use cases.  This line of research is dominated by works that address the data heterogeneity in the FL environment's sample space by altering the Federated Averaging (FedAvg) \cite{wahab2021federated} aggregation method that determines the weights of the shared model.  To that end, Probabilistic Federated Neural Matching (PFNM) \cite{yurochkin2018probabilistic}, Federated Matching (FedMA) \cite{wang2020federated}, FedProx \cite{li2020federated}, and FedNova \cite{fed_nova} contributed to the advancement of the state-of-the-art. However, this work targets the heterogeneity in the feature space, which renders the aforementioned methods unfit for the use case under study. 

Few works explored the model heterogeneity problem in the context of FL. A common approach combines transfer learning and knowledge distillation to address the non-uniformity of client models. In this approach, the local model weights that are transmitted to the global model are replaced by class outputs obtained using public data. These class outputs are obtained by training each client on public and private data. After that, local models are trained to approach the aggregated global model results. Methods such as FedMD \cite{li2019fedmd}, Cronus \cite{chang2019cronus}, FedDF \cite{lin2020ensemble}, and FedDistill \cite{jiang2020federated} follow this strategy with variations related to the knowledge distillation technique. Cronus \cite{chang2019cronus} incorporates the public and private datasets for training while FedMD \cite{li2019fedmd} discards the public data after the initial training phase. Cronus and FedMD apply the knowledge distillation on the local agents' side. To mitigate the effects of the changes to the public dataset on the performance of local clients, FedDF \cite{lin2020ensemble} performs the distillation on the server side using Generative Adversarial Networks (GANs). 

The last part discusses the literature that tackles the representational similarity of NNs, which mainly motivates our applied approach. SVCCA \cite{raghu2017svcca} and PWCCA \cite{morcos2018insights} are tools that compare two representations of NNs, which combine Canonical Correlation (CCA), and Singular Value Decomposition (SVD). On the one hand, the invariance to affine transformation promotes the use of CCA for comparing NN architectures. On the other hand, SVD determines the most important directions in the original data, which are fed to the CCA to compute the similarity in NN representations. 

The authors of \cite{li2015convergent, williams2021generalized, wang2020federated} discuss the permutation invariance of NN. This invariance suggests that NNs can create versions of the same architecture while permuting the parameters' order. This observation suggests the existence of drastic differences between neurons even when sharing the feature space and architectural design. The study by Li \textit{et al.} \cite{li2015convergent} uncovered the existence of one-to-one and one-to-many correspondence of neurons of the same architecture. This discovery validates the SVCCA approach, in terms of the existence of linear relationships between neurons, but, simultaneously, demonstrates that each NN uniquely creates its unique methods of feature engineering through neuron combinations. This fact shows that methods such as SVCCA do not fully capture the spectrum of similarity and new methods that mirror the non-linearity should be proposed. 

The surveyed literature exposes some limitations that motivate this manuscript. This paper is the first to address CO\textsubscript{2} predictions in a pre-defined time window application using the FL paradigm. Applications in smart buildings are confined to energy prediction, which demonstrates the novelty of this work and highlights its importance on both economic and health levels. The model heterogeneity approaches dominated by knowledge distillation methods necessitate the existence of public data, related to the problem, to establish common grounds between models of different architectures. To that end, generating a public dataset that resembles the private set can potentially violate the privacy-preservation aspect of FL. For these reasons, knowledge distillation approaches are limited to theoretical implementations, which undermines their utility in real-world scenarios. Mitigating the absence of models is neglected in the literature, which sheds light on this paper's contribution and insights into this critical problem in the FL paradigm and in the IoT environment in general. 

Inspired by the methodologies that explore the similarity between NNs and representational learning, our approach proposes CorrFL as a model aggregation method at the central server that mitigates the absence of models and generates highly correlated features. The direct implementation of the CorrNet on the studied use case is obstructed by multiple limitations. First, the original formulation and implementation consider only a use case consisting of two views. As such, it does not fit the multi-view nature of the IoT device models. Second, the loss function incorporates a weighted correlation between hidden representations to deduct from the sum of reconstruction losses. As a result, the correlation is an auxiliary factor, rather than a central one. To address these shortcomings, the formulations of the L1 and L2 losses are extended to multi-view representations to fit the diverse nature of IoT devices capturing environmental features. Moreover, the formulation of the correlation-based loss function is altered to maximize the correlation between hidden representations.

\begingroup
\setlength{\tabcolsep}{11pt} 
\renewcommand{\arraystretch}{1.5}
\begin{table}[]
    \centering
    \caption{Methodology Symbols}
    
    \begin{tabular}{c|c}
         \textbf{Symbol} & \textbf{Meaning}  \\
         \hline
         $L$ & Number of neurons in the first layer \\
         $n$ & Number of unique models \\
         $mi$ & Neural network models $i \in n$ \\
         $n_i$ & number of input features for models $mi$ \\
         $w_i$ & \makecell{$L \times n_i$ represents the number of weights \\ between the input layer and the first hidden layer} \\
         $\widehat{w_i}$ & restructured weights after applying CorrFL \\
         $L_1$ & reconstruction loss \\
         $L_2$ & reconstruction loss when one of the models is absent \\
         $L_3$ & correlation loss \\
         $W_i$ & set of model weights with an absent model $mi$ \\
         $h$ / $g$ / $\Psi$ & \makecell{encoding function / decoding \\ function / encoder-decoder} \\ 
        $H_i$ & \makecell{shared representation of the bottleneck \\ layer when a model $mi$ is missing} \\ 
        $r_i$ & \makecell{$r_i \in R$, where $R$ represents \\ the combinations of absent models' representation} \\
        \hline
    \end{tabular}
    \label{tab:set_methodology_symbols}
\end{table}
\endgroup

\section{Proposed Approach: Correlational Federated Learning}
\begin{figure*}
    \centering
    \includegraphics[scale=0.45]{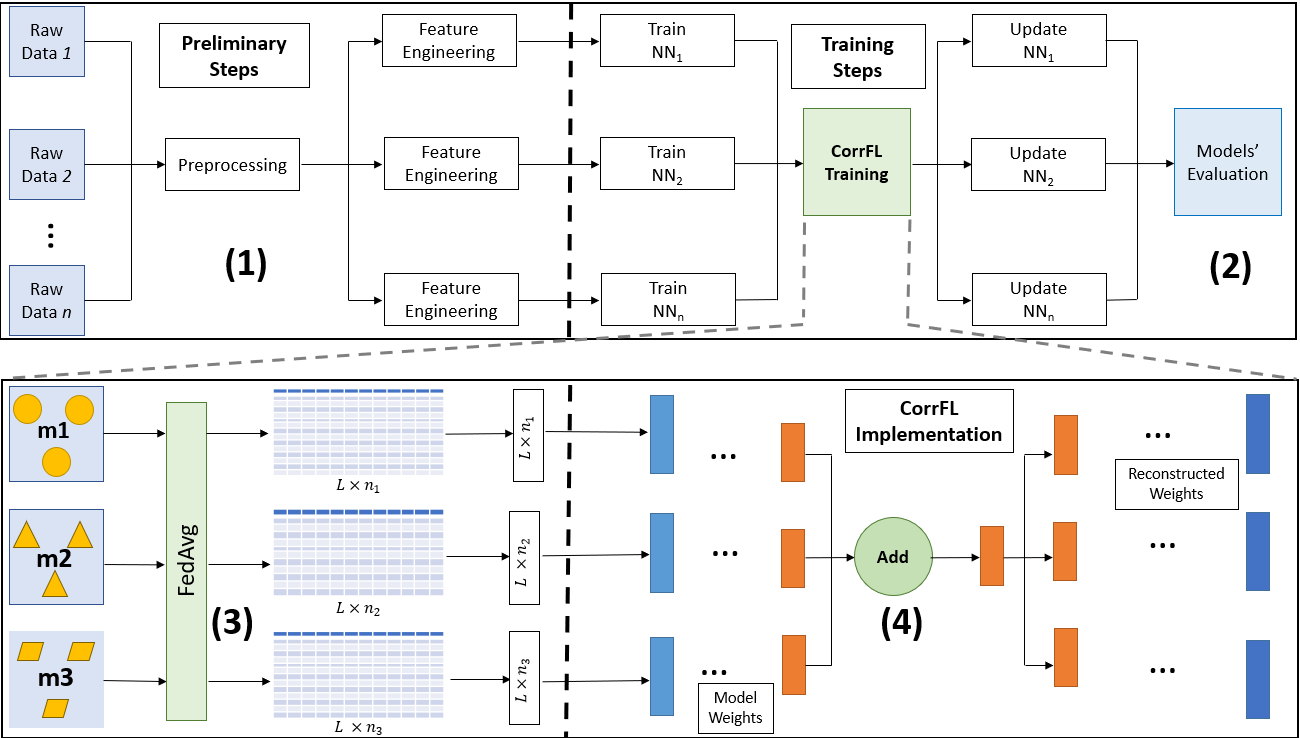}
    \caption{Components of CorrFL}
    \label{fig:methodology}
\end{figure*}

This section details how multi-view representational learning and correlational analysis are combined with the FL environment (CorrFL) to address model heterogeneity and availability constraints. The processes involved in the realization of the CorrFL are depicted in Figure \ref{fig:methodology}. The explanation focuses on the two subsystems that compose the FL paradigm: the local clients and the central server. The set of all symbols used throughout this section is summarized in Table \ref{tab:set_methodology_symbols}. 

\subsection{Local Client Training}
The environment under study encompasses different sets of IoT devices, each collecting environmental features to predict CO\textsubscript{2} concentrations. The FL paradigm commences with training local NNs on each IoT node in the environment. 
 This step is depicted in the left side of the dotted line of the upper part of Figure \ref{fig:methodology}, referred to by (1).  Each of the raw node data undergoes common preprocessing and feature engineering procedures. However, due to differences in the environmental features collected by each node, the feature engineering step produces different sets of input features. The differences in features post feature engineering step of each IoT are reflected in their corresponding NN architecture. Assuming common NN hidden layers, the input layer is the distinctive property of each local agent model, resulting in differences in the number of weights between the input layer and the hidden layer, and the input feature combinations in the subsequent hidden layers. The alteration of subsequent hidden layers will be addressed in future work. 

At the end of a pre-defined training time or a communication cycle, each client sends their NN weights to the central server.  The processes taking place in this phase are shown on the right side of the dotted line of the upper part of Figure \ref{fig:methodology}, referred to by (2).  At the central server, the CorrFL training begins, which represents the primary contribution of these works. The parameters sent by each node include the weights between the input layer and the hidden layer, the weights between hidden layers, and the weights between the last hidden layer and the output layer. 

\subsection{Central Server Aggregation}
 The applied processes in this step are shown in the lower part of Figure \ref{fig:methodology}.  The central server receives sets of model weights that need to be aggregated and transmitted back to the IoT devices when an unavailability concern arises. The devices are grouped based on their shared input layer, representing the features that resulted from the feature engineering step explained in the previous section. To demonstrate these differences, each model is depicted in Figure \ref{fig:methodology} with a specific shape, denoted by $m1$, $m2$, and $m3$. The central server's main objective is to mitigate the absence of models by generating them from the existing models. Two stages are implemented to realize this task.
 
The first stage of this procedure is analogous to a preprocessing step. It first aggregates the homogeneous models, that share the same feature space, which is achieved using the conventional FedAvg method. FedAvg calculates the element-wise average of the model's weights for each layer. This stage does not present any novelty compared to other approaches in the FL paradigm. Assuming that the number of neurons of the first hidden layer is $L$ for each model, the number of input features is $n_1$, $n_2$, and $n_3$ for models $m1$, $m2$, and $m3$, the resultant weight matrices are of size $L \times n_1$, $L \times n_2$, and $L \times n_3$, respectively. The matrices are later flattened to a 1-dimensional array and fed as inputs to the CorrFL.  These processes are depicted in part (3) of Figure \ref{fig:methodology}, representing the left side of the dotted line of the lower part of the figure. 

 The challenging aspect of this environment is combining the heterogeneous models while adhering to the central server's main objective. In the studied use case, the objectives can be summarized by the servers' need to generate models for IoT nodes that failed to dispatch their model updates due to connectivity issues. The adopted approach referred to as CorrFL extends CorrNet's architecture to multi-view representations and alters its loss function to fit the requirements and assumptions of the aggregated models and the environment's availability constraints.  A bird's eye view of the CorrFL implementation is depicted in part (4) of Figure \ref{fig:methodology}. 

In terms of its architecture, CorrFL extends the CorrNet architecture to multi-view representations that are fed to its input layer, keeping the remainder of its architectural structure intact. Here, input, model, and view are used interchangeably. CorrFL incorporates an AE for each view, which enables the projection of each model weights into a common subspace, also referred to as the latent space. The encoding function of the CorrFL is denoted by $h$ while the decoding function is referred to as $g$. Together, the encoding and decoding functions represent the CorrFL transformation functions denoted by $\Psi$. Defining CorrFL's loss function dictates the realization of the central server's objectives of mitigating the unavailability of local model updates.  To that end, the loss functions of CorrNet's original formulation, defined as L1 loss, L2 loss, and L3 loss, are extended and altered to meet the envisioned central server's objectives.



In the original formulation, the L1 loss ($L_1$) minimizes the differences between the original and the reconstructed model, when all the models are present. This loss does not directly contribute to the objectives set for the central server, but it alters the model weights so that the internal AE converges \cite{chandar2016correlational}. The staple reconstruction error is obtained using the mean square error (MSE) that amplifies large errors and diminishes the effect of smaller ones. Since this is the first work that applies AE in the field of model weight reconstruction, MSE is employed to calculate the reconstruction loss. However, a thorough investigation of the available loss functions should be carried out to find the loss function that fits the sensitive nature of the models' weights due to its effect on model performance. Assuming that $n$ models are present, whereby the weights of each input model are denoted by $w_i$ such that $i \in n$, the reconstructed weights are $\widehat{w_n}$, $L_1$ is defined as follows:
\begin{equation}
    L_1 = \frac{1}{n}\sum_{i=1}^{n} (w_i - \widehat{w_i}) ^ 2 
\end{equation}

The $L_1$ alone drives the CorrFL to reconstruct the inputs if they all exist; however, it does not address the models' absence. Therefore, this loss function should be augmented with other losses to fulfill the central server's main objective. The L2 loss ($L_2$) integrates the model's availability concern into its formulation. When all model weights are present, $L_2$ assumes that one model is not available, which reflects the studied IoT environment. In this scenario, $L_2$ calculates again the reconstruction loss. This step is crucial so that the models are properly reconstructed when one model is missing. Through the formulation of $L_2$, the common representation when a model is absent is obtained. This representation is retained to be used in the formulation of the L3 loss ($L_3$). The set of inputs, representing the set of weights is denoted by $W=\{w_1, ... w_i, ..., w_n\}$. The unavailability of a model $mi$ is emulated by setting its weights $w_i$ to be zeros. In this scenario, the set of inputs $W$ with missing weights $i$ is denoted by $W_i$. The formulation of $L_2$ is as follows: 
\begin{equation}
    L_2 = \frac{1}{n} \sum_{i=1} ^ {n} (\Psi(W_i) - w_i)
\end{equation}

The $L_2$ provides, as a by-product, the models' common representation when a model is missing. For $W_i$ input, the common representation $H_i$ is obtained as follows: 
\begin{equation}
    H_i = \sum_{i=1}^{n} h(w_i)
\end{equation}

The data heterogeneity and the intersection of the models' feature spaces produce NN models that can potentially exhibit high correlations in some weights. An option to quantify this correlation is conducting the one-to-one or the one-to-many correspondence on the hidden layers \cite{li2015convergent}. However, this approach is computationally expensive due to a large number of input weights and the underlying assumption of linearity. Therefore, a correlation-based method that incorporates non-linearity is favourable. The non-linearity is achieved using the AE architecture that maps the input with different dimensions into a common latent space. When a model is absent, it is desired to construct this model from other models, knowing that this unavailable model should incorporate some aspects of the available ones. This requirement can be achieved by maximizing the correlation between common representations when a model is absent ($H_i$) reflected by  $L_3$. Here, the definition of $L_3$ diverges from the L3 loss definition in the original formulation. In particular, the L3 loss acts as an auxiliary loss that is deducted by a specific factor from L1 and L2 losses. This formulation suggests that it is nice to have property instead of a fundamental one. Therefore, to restore parity between different losses and incentivize the production of highly correlated models, a new definition of $L_3$ is proposed. To obtain $L_3$, the correlation between each pair of hidden representations is calculated. The combinations of all these representations are denoted by $R$. The pairwise correlation denoted by $c(r_i)$, such that $r_i \in R$, whereby $r_i$ is equivalent to the combination of $i_1$ and $i_2$ ($i_1 \neq i_2$) is calculated as follows:

\begin{gather}
    c(r_i) = \frac{(h(w_{i_1}) - \overline{h(w_{i_1})}) \times (h(w_{i_2}) - \overline{h(w_{i_2})})}{\sqrt{ (h(w_{i_1}) - \overline{h(w_{i_1})}) ^ 2 \times (h(w_{i_2}) - \overline{h(w_{i_2})}) ^ 2 }} \\
    L_3 = \sum_{r_i \in R} 1 - c(r_i)
\end{gather}

This $L_3$ is an altered version of CorrNet's original L3 loss formulation, which only deducts $c(r_i)$ from L1 and L2 losses. This new version of $L_3$ increases when the correlation is either low or negative and decreases otherwise. This definition drives the production of highly correlated reconstructed model weights, which are in agreement with the heterogeneous environment that typically results in highly correlated model weights. After the formulation of the $L_1$, $L_2$, and $L_3$, the loss function $L$ that directs the CorrFL model is as follows: 
\begin{equation}
    L = L_1 + L_2 + L_3
\end{equation}

\section{Experimental Procedure}
This section explains the experimental procedure that discusses the dataset used and its processing steps, feature engineering steps, neural network architecture, testing parameters, and implementation details.

\begingroup
\setlength{\tabcolsep}{5pt} 
\renewcommand{\arraystretch}{1.3}
\begin{table}[]
\centering
\caption{Feature Distribution among IoT Nodes}
\begin{tabular}{|c|c|c|}
\hline
\textbf{Model Name} & \textbf{IoT Nodes} & \textbf{Collected Features} \\ \hline
\multirow{4}{*}{m1} & node\_913 & 
\multirow{4}{*}{\{humidity, temperature, pressure, activity\}} \\ \cline{2-2}
 & node\_914 &  \\ \cline{2-2}
 & node\_915 &  \\ \cline{2-2}
 & node\_916 &  \\ \hline
m2 & node\_920 & \{humidity, temperature, pressure\} \\ \hline
m3 & node\_924 & \{humidity, temperature, pressure, CO\textsubscript{2}\} \\ \hline
\end{tabular}
\label{tab:dataset}
\end{table}
\endgroup

\subsection{Dataset Insights and Preprocessing}
The dataset includes environmental features collected over a year in the Nordic climate of Northern Finland using a host of sensors. Only six sensors are utilized to illustrate the utility of the proposed approach. The set of environmental features captured by each sensor is summarized in Table \ref{tab:dataset}. The activity level is calculated by aggregating the movement levels in each five-second interval for a one-minute granularity. Other features are captured with one-minute granularity. The dataset is collected in 2 conference rooms that can fit 12 people and 11 cubicles that can fit 2 people.  The CorrFL methodology is implemented in one of the conference rooms that fit 12 people, referred to as room00. For this study, the data corresponding to each sensor was first sorted based on its timestamp \cite{dataset}. Then, the data are aligned to start at the same timestamp and then sampled with one-minute granularity. Lastly, the gaps in data caused by communication issues are mitigated by interpolating the missing data using each feature's median. This way, with every communication round, the client models are trained using the same amount of data so no weighted approach is applied.

\subsection{Feature Engineering and Neural Network Architecture}
The original work that made the dataset available for this study explored different supervised ML techniques, including NN models, to predict the CO\textsubscript{2} concentrations in a future time horizon using data of the history time window \cite{dataset_work}.  Since only $m3$ type sensors involve the collection of CO\textsubscript{2} concentrations, the collected values are shared between all participants. This assumption follows the trend in vertical FL \cite{wahab2021federated}, whereby all the local agents own the labels, in this case, the CO\textsubscript{2} concentrations.
The paper that made available the dataset used throughout this manuscript has experimented with different feature combinations, with the goal of enhancing the CO\textsubscript{2} predictions over a future time horizon \cite{dataset_work}. The combination of lagged versions of environmental features over the history time window in addition to the difference in values of each feature between the end and the start of the history time produced satisfactory results. This feature engineering step incorporates time dependencies into the NN models \cite{shaer2022sound, CO2_MDPI}.  

The input layer size specific to each sensor differs depending on the collected features. One hidden layer with 16 neurons augments the input layer. This layer is connected to a single output layer that predicts the CO\textsubscript{2} concentrations. As highlighted in the methodology section, the input layer is the source of model heterogeneity as the other layers are common between different models. This architecture was chosen because it produced satisfactory results in the original study, and it is a good starting point to benchmark the CorrFL to be later extended to more pervasive and heterogeneous architectures. A single history and future time window of 5 minutes to serve as proof of concept for CorrFL. 

\begingroup
\setlength{\tabcolsep}{11pt} 
\renewcommand{\arraystretch}{1.5}
\begin{table}[]
    \centering
    \caption{Testing Parameters}
    \begin{tabular}{c|c|c}
         \textbf{Symbol} & \textbf{Meaning} & \textbf{Values}  \\
         \hline
         $CC$ & Communication cycles & 15 per epoch \\
         $d$ & delay & $\{1, 5, 10, 15\}$ \\
         $MDF$ & Model Dispatch Frequency &  $\{5, 10, 15\}$ \\
         $VE$ & Validation Epochs & $1 \rightarrow{10}$ \\
        \hline
    \end{tabular}
    \label{tab:testing_parameters}
\end{table}
\endgroup

\subsection{Testing Parameters}
The explanation of the considered parameters is divided based on the sub-systems of the FL environment. On the client side, many parameters related to the environment and the NN can be optimized. NN's performance can be improved by applying hyper-parameter optimization, encompassing the number of hidden layers, the activation functions, and the learning rate \cite{yang2020hyperparameter}. This process is deemed unessential to the primary purpose of the methodology, and it is left for future work or keen practitioners to explore. 

The more important concerns pertain to the communication cycles ($CC$) and the number of model weights sent at once, referred to as Model Dispatch Frequency ($MDF$), with each $CC$. These parameters determine the amount of data fed to the CorrFL model and dictate its convergence. Accordingly, a prolonged grid search is required to find the optimal combinations of these parameters. After applying some filtration on outlier instances of environmental features, the remainder of the dataset includes 381,419 data points which translates to around 265 days. To reduce the parameter search space, each $CC$ is assumed to be equal to 14 days, which means that local models train on 20,160 data points, before dispatching their accumulated model weights. The batch size of the training data is assumed to be 8. Investigating the effect of $CC$ and the batch size is left for future work. CorrFL evaluation experiments with $MDF$ equal to 5, 10, and 15, which means that model weights are collected at every 5th, 10th, or 15th batch. The time equivalent of these values is a model dispatch every 40, 80, and 120 minutes. The delay $d$ between the start of the training process on the client side and the server side is also a salient concern. The parameter $d$ here dictates the amount of training data accumulated to train the data-intensive CorrFL architecture and to stabilize the local agents' training. The delay $d$ represents some multiplier of $cc$. $W_{cc}$ is the set of weights at communication cycle $cc_i$ such that $cc_i \in CC$. The input fed to CorrFL is as follows:
\begin{equation}
    \bigcup_{i=1}^{d} W_{cc_{i}}
\end{equation}

While the research community has investigated the stability of NN learning \cite{forti1994necessary, ji2015further}, a consensus is yet to be reached about concrete analytical methods that quantify the time when this stability is attained. Therefore, in this experimental procedure, empirical evidence is key to identifying the point of relative stability that determines the $d$ parameter. Each NN is initialized with a random set of weights, obtained from the central server. Therefore, a reasonable $d$ starts with at least the first epoch. During the first epoch, the local NNs learn different facets of the training data, allowing the weights to be altered to match the data trend. As a result, each local model exits the randomness of weight initialization to weights that are reflective of the underlying heterogeneous data distributions. While sending the weight updates after the first learning epoch is based on theoretical assumptions, delaying this process beyond that point is predicated on empirical evidence. Analyzing the effect of $d$ on model performance is crucial to determine the utility of CorrFL and its capability to produce good results under different environments. Therefore, delays of one, five, ten, and fifteen epochs are assumed, equivalent to $15$, $75$, $150$, and $225$ $CC$. As for the CorrFL architecture, a deep encoder/decoder of two hidden layers that each include 128 and 32 neurons with no activation functions.

The developed NN models and evaluation criteria were built using PyTorch \cite{NEURIPS2019_9015} python libraries. The developed models were evaluated on Windows 10 PC with a 3.00 GHz 24-Core AMD Threadripper processor, 128 GB of RAM, and 8 GB Nvidia GeForce RTX 3060 Ti GPU. The code is made available on the GitHub repository\footnote{https://github.com/Western-OC2-Lab/CorrFL}. 

\section{Use Case and Evaluation Criteria}
This section describes the use case to evaluate the CorrFL. 
It first explains the environment under study, the motivations for applying this scenario, how this scenario transpires, and the evaluation criteria. 

\begin{figure}
    \centering
    \includegraphics[scale=0.26]{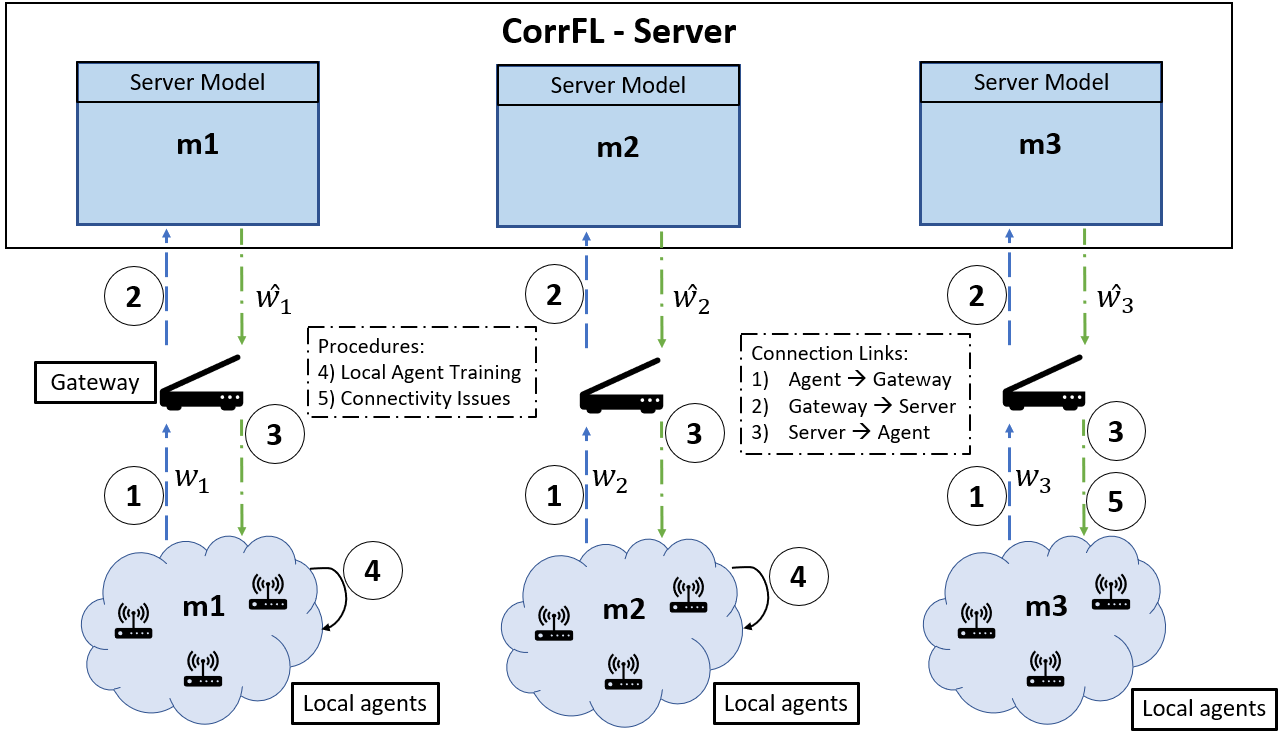}
    \caption{Bird's Eye View of the Use Case}
    \label{fig:Architecture}
\end{figure}

\subsection{Environment Description}
Before delving into the details of the adopted use case to evaluate the CorrFL, it is important to provide a bird's-eye view into the interactions occurring between the local models and the Server applying the CorrFL approach. Figure \ref{fig:Architecture} depicts the studied environment and follows the notation established throughout the paper. The sets of IoT devices are grouped based on their collected environmental features. In the illustration, these groups are referred to according to their corresponding local model weights $m1$, $m2$, and $m3$. 

Figure \ref{fig:Architecture} shows three connectivity links that can experience issues that affect the FL processes and jeopardize the availability of each set of sensors. The first link, denoted by 1, connects the set of IoT nodes to a gateway. The second link, denoted by 2, connects the gateway to the central server. Links 1 and 2 are referred to as the uplinks. On the other hand, the links that connect the server to the gateway and the gateway to each set of nodes are denoted by 3 and referred to as the downlink. In this paper, connectivity issues are encountered on links 1 and 2, whereas the downlink functions normally. The cut in connectivity through link 1 or link 2 is assumed to be long enough so that the ambient environment undergoes a major shift in its properties and underlying relationships. Use cases whereby intermittent connectivity issues are encountered are out of the scope of this paper. 

\subsection{Motivation}

The CorrFL approach is evaluated when two events transpire. The first event refers to the occurrence of connectivity issues for one of the model sets. The second event takes place after this disconnection when the characteristics of the environment, described by one or more environmental feature, change drastically. As such, the CorrFL will be tasked with producing model weights of the unavailable IoT devices from the updated models of the available ones. However, it is important to propose a plausible scenario relevant to the studied office environment and its effect on CO\textsubscript{2} predictions. 
 
Measuring activity levels can act as a proxy indicator of occupancy, which has shown a strong association with the variation of CO\textsubscript{2} concentrations \cite{eini2021smart}. In a large conference room that fits 12 people, radical changes can be experienced in occupancy. Since these rooms are dedicated to large meetings that are rarely conducted, it is expected that these rooms are left unoccupied most of the time. Team and executive meetings are rare events, which further enforce the premise of the domination of low occupancy events. Moreover, the emergence of remote work facilitated by the recent pandemic has accentuated this trend \cite{motuziene2022office}, which degraded the occupancy prediction models developed before the pandemic. 

\begin{figure*}
    \centering
    \includegraphics[scale=0.23]{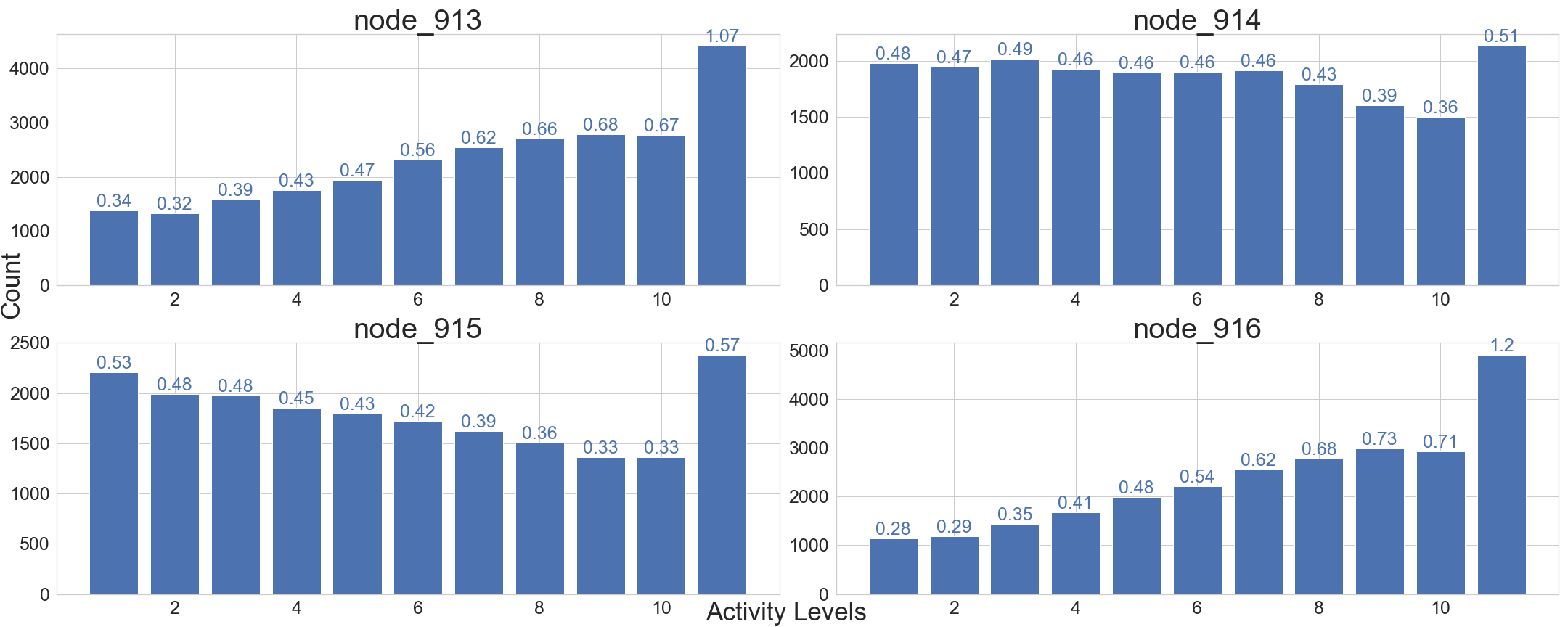}
    \caption{Activity Level Distribution between Sensors}
    \label{fig:pir_distribution}
\end{figure*}

This trend of low occupancy is also observed in the dataset used in this paper. To showcase this trend, Figure \ref{fig:pir_distribution} depicts the distribution of activity levels greater than 0 over the subset of sensors capturing this environmental feature. The percentage of data with the corresponding activity level is shown on top of each bar. As expected, the dataset is dominated by instances with no activity levels, as the percentage of data with any activity level is in the range of 5-8\%. Moreover, extreme activity levels are more frequent in such an environment. The upper half of the existence of activity levels, in the range of 8-12, outnumber the lower levels. This observation aligns with the main function of the conference room, which is only used for larger meetings. The combination of extracted observations and the post-pandemic office environment engenders the perfect recipe for evaluating the CorrFL approach. For this manuscript, the adopted occupancy use case is considered for activity levels that are above 7. This threshold perfectly balances establishing a rare event and supplying enough data for any ML model. Since each sensor monitors a specific aspect of the environment, it is unlikely that the individual sensors would capture the same activity levels in their vicinity. Therefore, the common timestamps, when the studied use case condition is fulfilled are rare if any, which puts any ML model at a disadvantage. As a result, the union of these timestamps is considered, which provides more data, but would include some aspects of previous data characteristics that are detached from the adopted use case. Overall, the total number of timestamps resulting from this process constitutes 15\% of the total amount of data, enough for 3 $CC$. 

\subsection{Use Case in Action}
The data is split into a training dataset and a testing dataset. The training dataset represents the data under normal conditions, whereby the activity levels are under the pre-defined threshold occupancy levels. On the other hand, the testing dataset includes the dataset with occupancy conditions corresponding to activity levels that are above 7. In the training phase, each local agent is trained using its local data and ML models. Meanwhile, the CorrFL model is progressively trained by the weights generated by each local model, depending on the $MDF$ parameter. During the training process, it is assumed that no availability issues are experienced and all links 1, 2, and 3 are fully functioning. 

\begin{figure}
    \centering
    \includegraphics[scale=0.25]{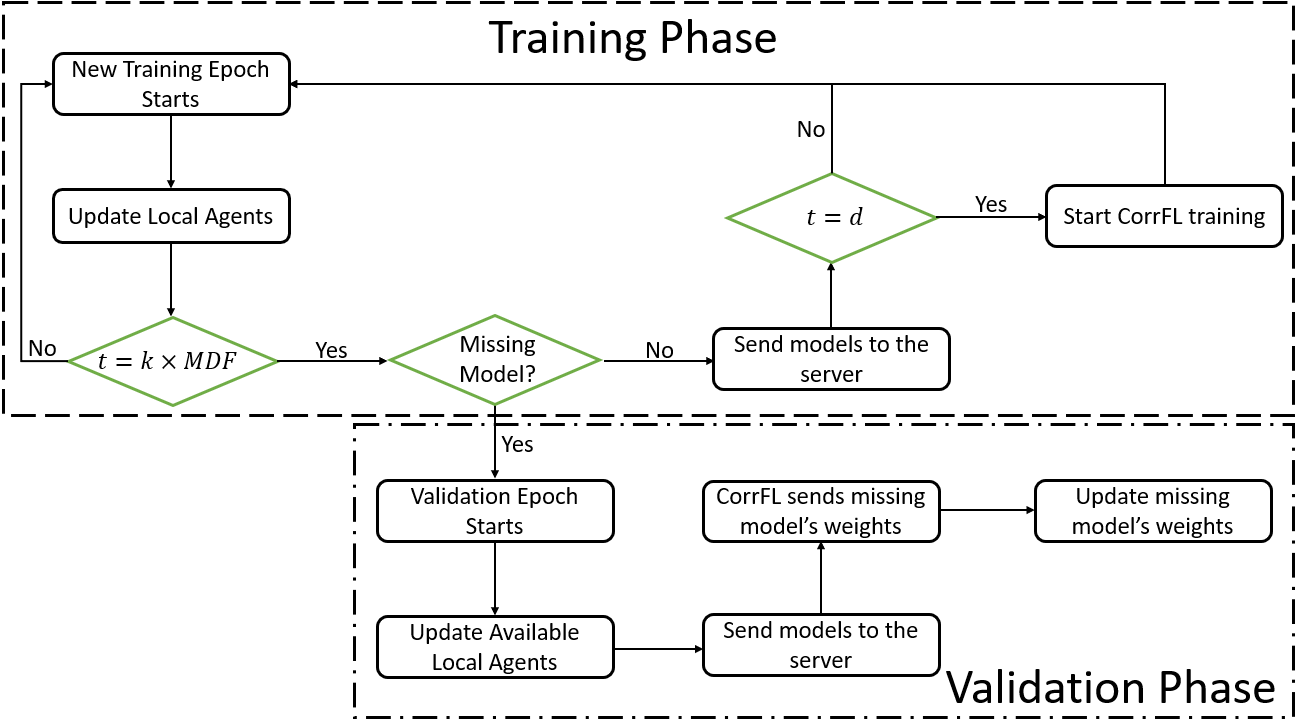}
    \caption{CorrFL Flowchart}
    \label{fig:corrFL_flowchart}
\end{figure}

CorrFL's evaluation process begins when one set of IoT devices is absent, caused by the disconnection of links 1 or 2. In this manuscript, set $m3$ is assumed to experience uplink connection downtime but still able to send the new CO\textsubscript{2} concentrations required for the predictions of other models. Here, the environment flips into heightened activity-level conditions. The testing set is split into validation and testing sets. Each of the available sets of IoT devices, denoted by  $m1$ and $m2$, continue training on their respective validation sets. This process is denoted by 4 in Figure \ref{fig:Architecture}. The number of times the available models train on the validation set before the CorrFL sends the updated weights $\widehat{w_3}$, denoted by 5 in Figure \ref{fig:Architecture}, is referred to as the validation epoch ($VE$), which represents an additional parameter to experiment with.  The $VE$ can also represent the number of training cycles that the model $m3$ is missing and CorrFL is trying to compensate. Therefore, the analysis of the effect of $VE$ translates to the effect of a model being missed over multiple training cycles. 
The $\widehat{w_3}$ representing the $m3$'s first layer generated weights are combined with $m3$'s second layer weight obtained solely from the training set. Together, the newly created model is referred to as a CorrFL model or $m3\_CorrFL$.  During the validation process, the CorrFL uses its pre-trained models to generate the missing model and no training of CorrFL is executed. During the testing phase, the available local models only predict the CO\textsubscript{2} concentrations and no training process is involved on their end. The testing parameters are summarized in Table \ref{tab:testing_parameters}.  Figure \ref{fig:corrFL_flowchart} summarizes the steps involved in the training and validation phases of CorrFL. 

\subsection{Evaluation Criteria}
The evaluation process is split into three phases, training phase, validation phase, and testing phase. Regardless of the ongoing phase, the developed models' predictions are evaluated using the Mean Absolute Error (MAE).  It is important to establish the connection between the MAE metric and its contribution to the reduction of HVAC energy consumption. The prediction of CO\textsubscript{2} concentration is instrumental in the activation of HVAC systems. While small deviations in predictions are insignificant in the short term, they can contribute to the superfluous activation of HVAC systems, increasing their overall energy consumption. As such, the envisioned application values small and large deviations equally, which favours the employment of MAE to measure the accuracy of CO\textsubscript{2} predictions. During the validation phase, the CorrFL generates an updated model $\widehat{w_3}$ for the set $m3$ of unavailable IoT devices.  This model is compared to the model trained only on the training set, referred to as $m3\_benchmark$. MAE in CO\textsubscript{2} predictions is the basis for this comparison. 

The second evaluation criterion encompasses networking concerns pertaining to the FL environment. The additional exchange of models during the validation process presents a considerable communication burden on the networking system. Therefore, to quantify this burden, the number of Megabytes (Mbs) per percentage of improvement (PI) between the CorrFL approach, and the benchmark model weights is calculated. The total memory usage of the CorrFL during the validation phase is denoted by $U$. The MAE of CO\textsubscript{2} predictions of the CorrFL approach is denoted by $p_{C}$ and the MAE of CO\textsubscript{2} predictions of the benchmark model is denoted by $p_{B}$. The improvement ratio is $IR$ and the PI formulations are as follows:
\begin{gather}
    IR = 100 \times \frac{ (p_{C} - p_{B}) }{p_{B}} \\
    PI = \frac{U}{IR}
    \label{eqn:PI}
\end{gather}

\section{Results}
This section provides a detailed analysis and discussion of the CorrFL performance under different experimental parameters and draws conclusions and suggests future directions. 

\subsection{CorrFL Evaluation}
This section discusses the general results of applying the CorrFL approach to the explained scenario. The generality refers to simply averaging the $m3\_CorrFL$ model over all configurations, including the $d$, $VE$, and the $MDF$. This comparison allows extracting the general trend of CorrFL's approach, regardless of the underlying assumptions about the convergence of local models.

\begin{figure}
    \centering
    \includegraphics[scale=0.17]{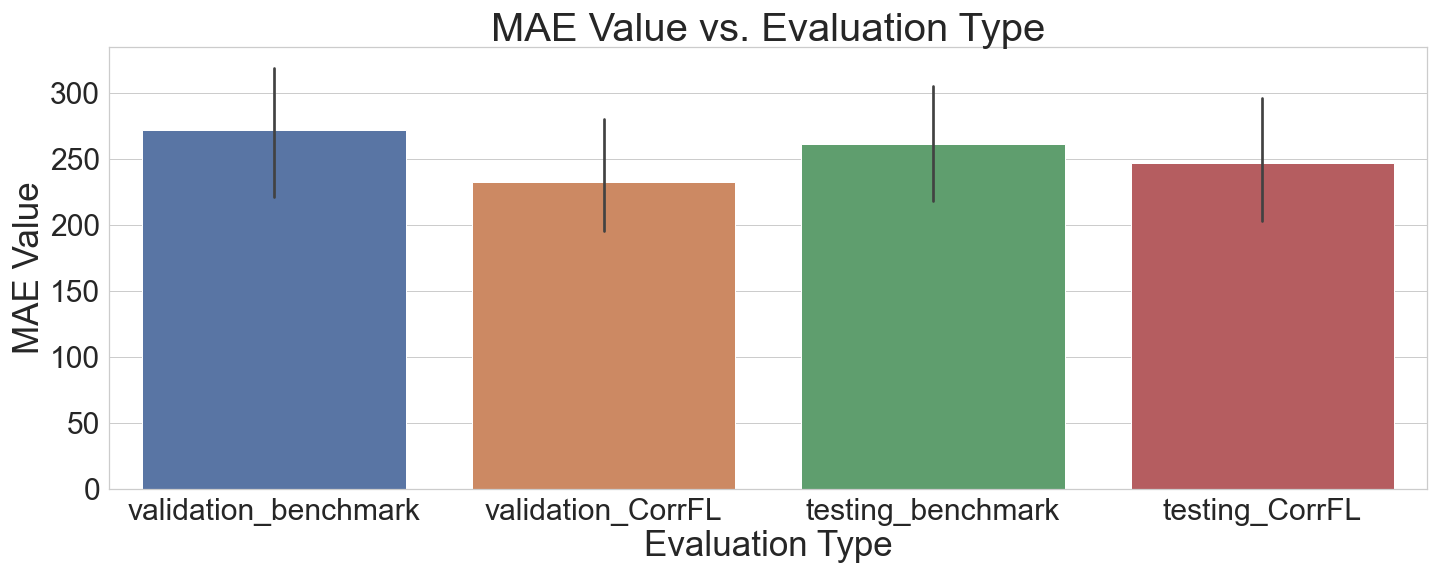}
    \caption{Effect of CorrFL}
    \label{fig:general_corrFL}
\end{figure}

Figure \ref{fig:general_corrFL} shows the average results of the CorrFL vs. the benchmark model for validation and testing datasets. The inclusion of the validation set for evaluating the CorrFL approach assesses the available local model's ability to incorporate aspects of the high activity level use case into their models. This capability is reflected by altering their weights so that the weights yielded for the absent model are generalized over the whole testing set. 

With regard to the benchmark model, no stark differences exist in the average and the standard deviation of its performance on the validation and the testing set with a slight edge for the testing set. Despite the uniform method adopted to split the validation and testing sets, these results show that the testing set has more common properties with the training set than the validation set. The observations on the benchmark model are reversed for the CorrFL models. In particular, the generated models perform on the validation set better than on the test set. This advantage is expected because the available models are trained on the validation set, which yields model weights that are better adjusted to this set. Additionally, the worse testing results align with the observations extracted in relation to the benchmark models. The training of the available models on the validation set incorporated some of its aspects in the updated model weights, which suggests that previously learned environment dynamics are gradually being replaced. This fact combined with the preposition of the existence of some training data aspects in the testing set explains the results obtained by the CorrFL models on the testing set. 

This section alludes to the superiority of the CorrFL models over the benchmark models for the heightened activity level use case. However, the effects of different configuration parameters on the convergence of local models and the quality of CorrFL models are concealed by only reporting the average MAE. Moreover, the relatively large standard deviations show that there are more interesting insights about the performance of the generated and benchmark models. The PI criterion is not included in this section because it depends only on the validation dataset. Analyzing these parameters is of paramount importance and concluding their effect opens many research questions for keen practitioners to answer. 

\subsection{Effect of delay}
This subsection details the effect of the delay parameter on the MAE and PI of the $m3\_CorrFL$ in comparison with the $m3\_benchmark$. Since the number of collected models and the $VE$ contribute to performance variations, the analysis that follows alters $d$, keeping other parameters the same. This subsection begins by first discussing the effect of $d$ on the performance of the local models and as a result its contribution to $m3\_CorrFL$ performance.

\begin{figure}
    \centering
    \includegraphics[scale=0.19]{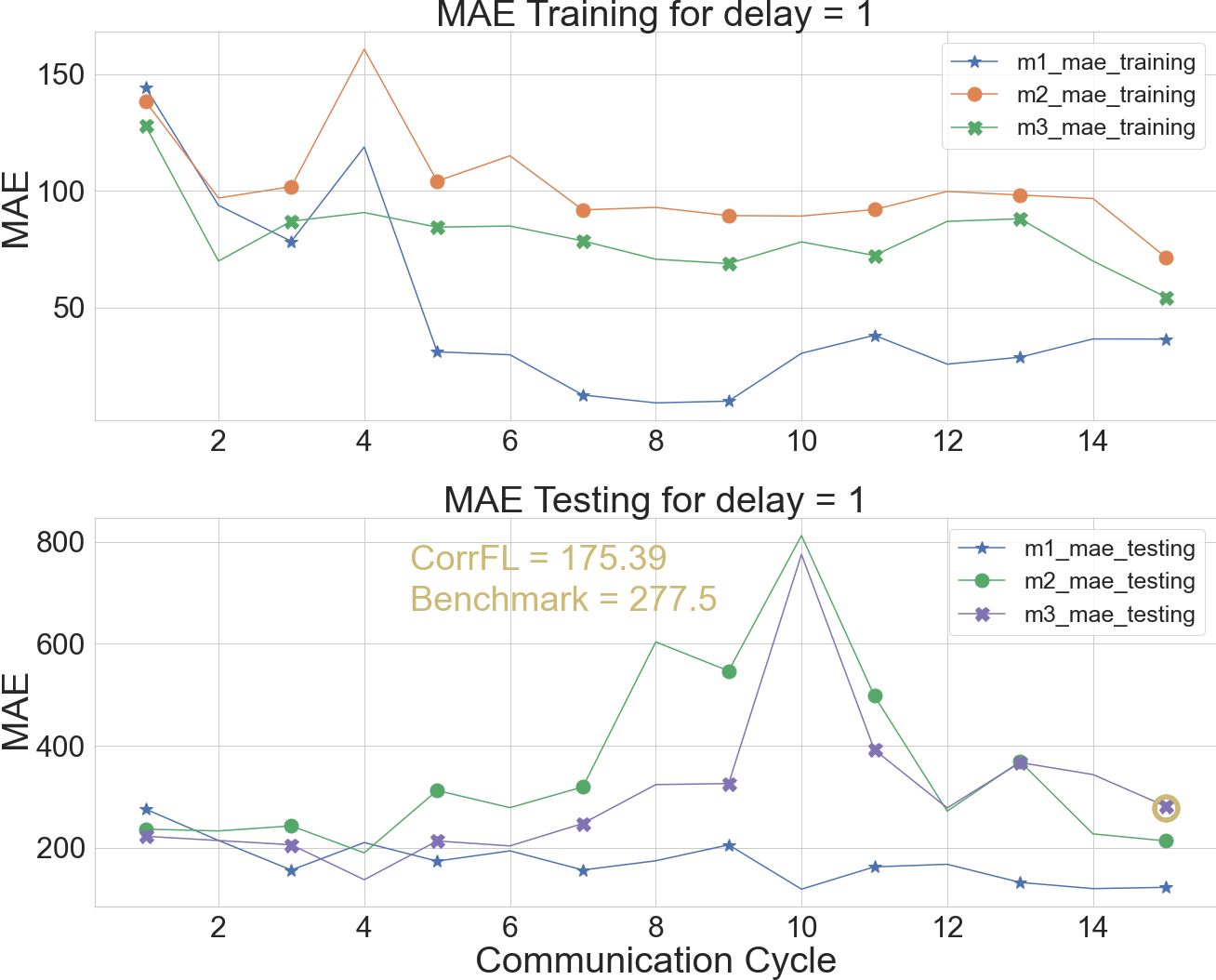}
    \caption{MAE training and testing for Delay = 1}
    \label{fig:train_test_15}
\end{figure}

\begin{figure}
    \centering
    \includegraphics[scale=0.19]{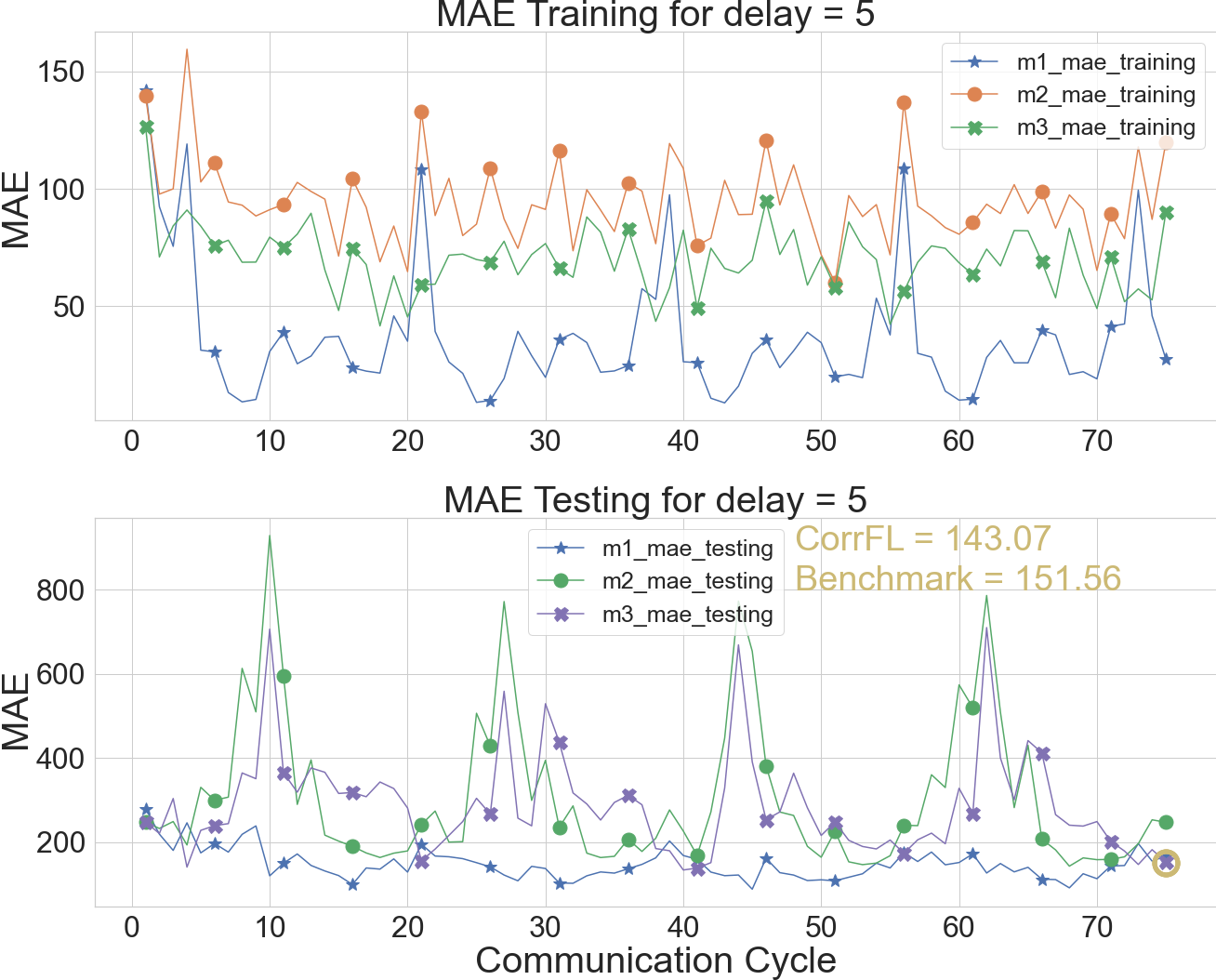}
    \caption{MAE training and testing for Delay = 5}
    \label{fig:train_test_75}
\end{figure}

\begin{figure}
    \centering
    \includegraphics[scale=0.19]{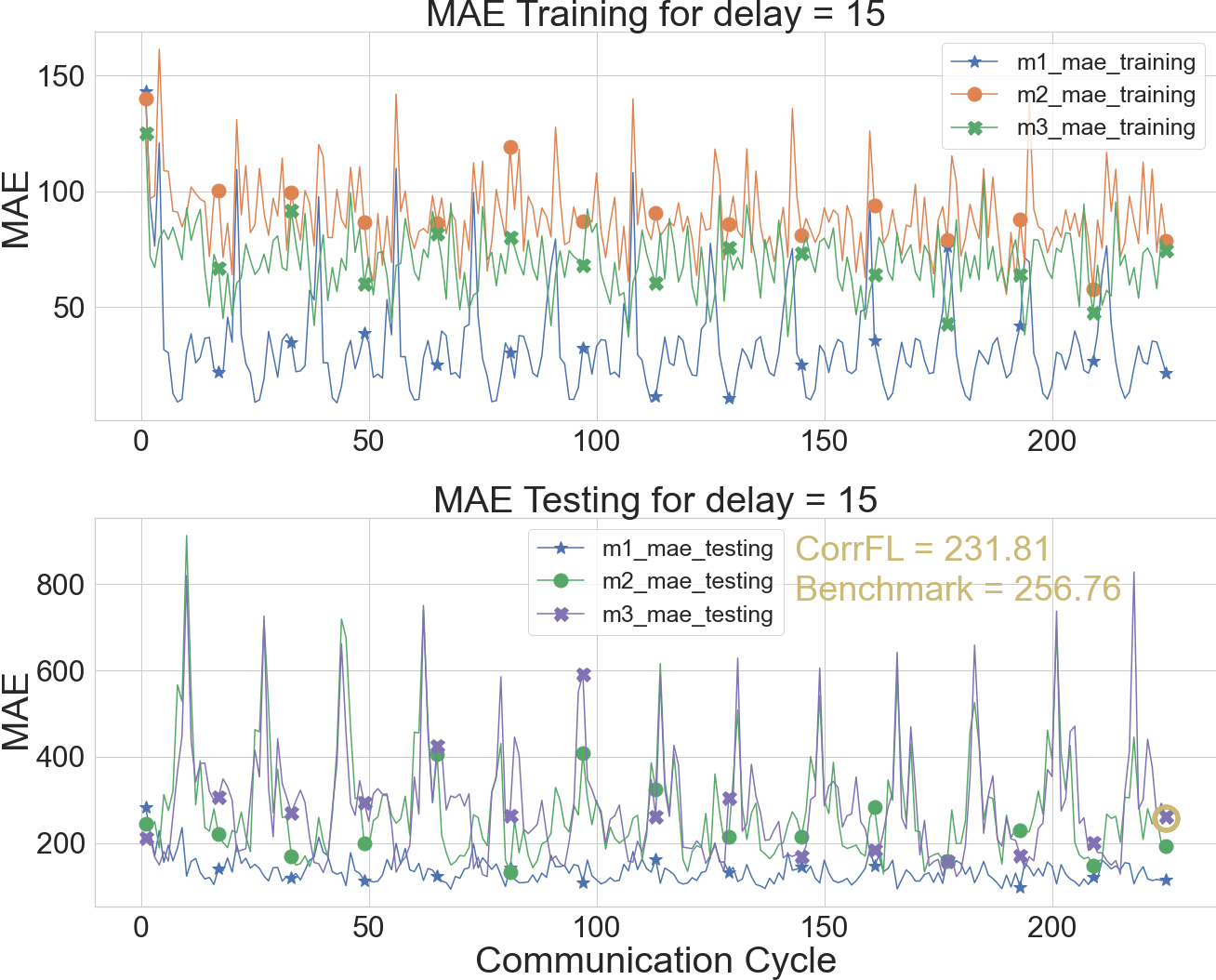}
    \caption{MAE training and testing for Delay = 15}
    \label{fig:train_test_225}
\end{figure}

Figures \ref{fig:train_test_15}, \ref{fig:train_test_75}, and \ref{fig:train_test_225} show the variations for MAE training and testing with the delay, considering a $VE=1$ and $MDF=5$. The IoT nodes are grouped as per the description and notation introduced in Table \ref{tab:dataset}. The performance of the benchmark model on the validation set is highlighted using a circle in Figures \ref{fig:train_test_15}, \ref{fig:train_test_75}, and \ref{fig:train_test_225}. Moreover, each figure of MAE testing includes the performance results of the benchmark model and the models generated by CorrFL. 

Each of the Figures \ref{fig:train_test_15} and \ref{fig:train_test_75} represents a microcosm of Figure \ref{fig:train_test_225}. For example, the MAE training experiences spikes in its value on $CC = 4$ for $d = 1$ and models $m1$ and $m3$. The same pattern is observed for the $d = 5$ and $d = 15$, which implies that the training process is consistent with different delays. As for the differences between models $m1$, $m2$, and $m3$, the $m3$ significantly outperforms other models on its training set and on the test set in the extreme delay conditions of $d = 1$ or $d = 15$. The differences are less prominent for the delay $d = 5$, such that $m3$ outperforms $m1$. In general, these results rank the importance of some features in the CO\textsubscript{2} predictions. The model $m2$, which includes features representing the least common denominator among all the available features performs relatively poorly compared to other models. On the other hand, the best-performing models $m1$ include activity levels, which shows the importance of the occupancy's inclusion in predicting the CO\textsubscript{2} concentrations. 

\begingroup
\setlength{\tabcolsep}{6pt} 
\renewcommand{\arraystretch}{1.2}
\begin{table}[]
\centering
\caption{PI for different Delays}
\begin{tabular}{|c|c|}
\hline
\textbf{Delay (epochs)} & \textbf{PI (Mb)} \\ \hline
1              & 0.12    \\ \hline
5              & 0.81    \\ \hline
15             & 0.47    \\ \hline
\end{tabular}
\label{tab:PI_train_test}
\end{table}
\endgroup

In all the considered delays, $m3\_CorrFL$ models have outperformed the $m3\_benchmark$ models. The starkest differences in performance are reported for $d = 1$. This observation suggests that $m3$ is yet to converge on the training set and integrate some of its characteristics. On the other hand, the CorrFL model employing the relatively well-performing $m1$ and $m2$ models is superior to the benchmark model. The performance gap significantly shrinks with $d = 5$. With the addition of more epochs, all models improved their MAE, contributing to generating better-performing CorrFL models compared to $d=1$. This observation also applies to $m3$ that outperforms its counterpart for $d=1$. Beyond $d = 5$, the MAE testing of the benchmark and the CorrFL models significantly plummets. These poor results can be attributed to several factors. On the $m3\_benchmark$'s model end, extending the training process to include more epochs overfits the model on the training data that possesses characteristics that are significantly different from the testing set. On the $m3\_CorrFL$ model's end, the degradation of performance can be either attributed to the overfitting of $m3$, which is established through previous observations, or the overfitting of all models. The former reason indicates the profound contribution of $m3$ in the training process of CorrFL. In either of the cases, the validation process conducted by $m1$ and $m2$ contributed to enhancing the performance of the CorrFL model.

Table \ref{tab:PI_train_test} summarizes the PI metric as a result of the different $d$. Since a single $VE$ is considered, the number of models dispatched to the central server in that period is 778 models. Applying equation \ref{eqn:PI}, the obtained results show that the utility of the $m3\_CorrFL$ model diminishes when the $m3\_benchmark$ performs well compared to the $m3\_CorrFL$ model. This observation is evident for $d = 5$ when small gains in performance are attained. The greatest advantage of employing CorrFL is acquired at the very first epoch, due to the large performance gap between the two models. 

\begin{figure*}
    \centering
    \begin{subfigure}[t]{0.45\textwidth}
        \centering
        \includegraphics[scale=0.22]{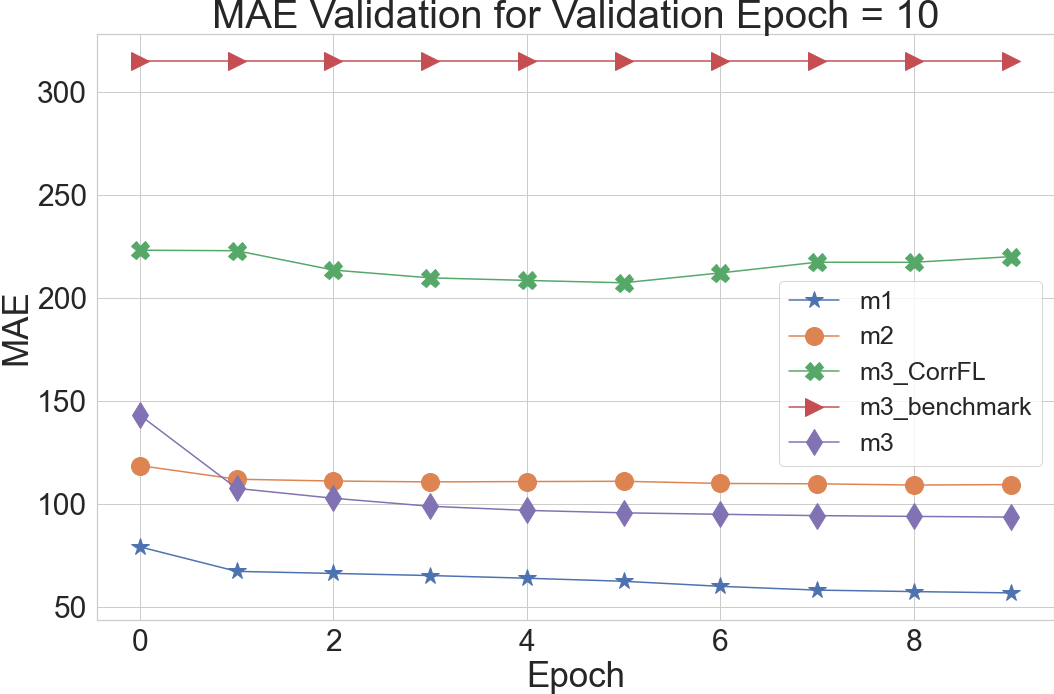}
        \caption{10 $VE$ for 10 $d$}
        \label{fig:10_VE_150_D}
    \end{subfigure}%
    ~ 
    \begin{subfigure}[t]{0.45\textwidth}
        \centering
        \includegraphics[scale=0.22]{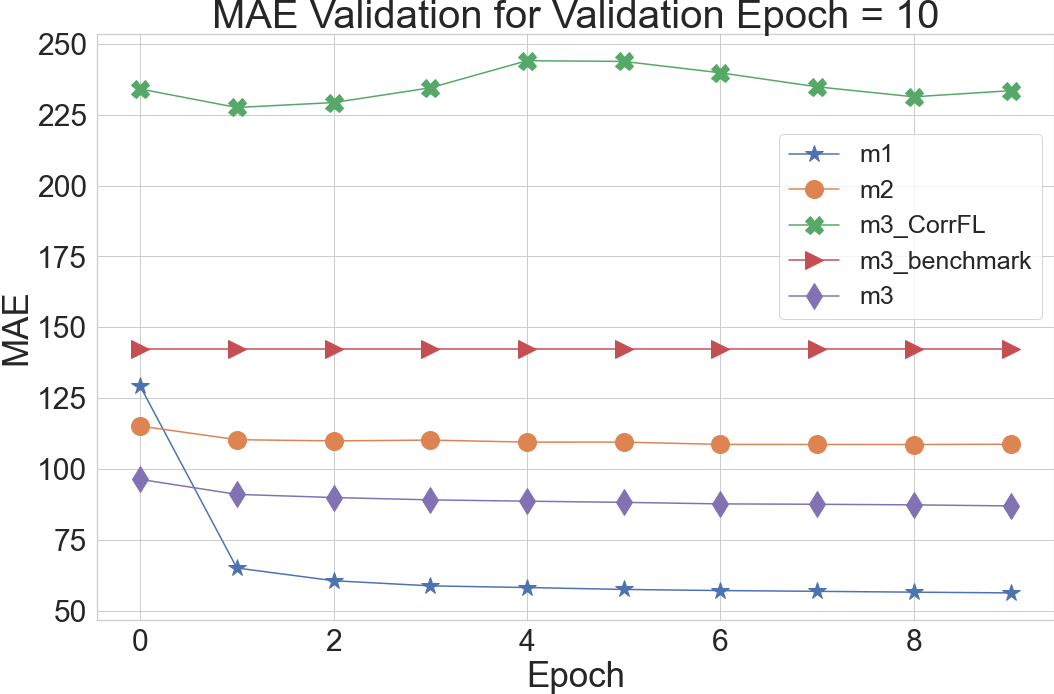}
        \caption{10 $VE$ for 5 $d$}
        \label{fig:10_VE_75_D}
    \end{subfigure}
    \caption{Variation of MAE Validation with different Validation Epochs}
\end{figure*}

This analysis emphasizes the importance of $d$ in the CorrFL models' performance, and the significance of local agents' convergence on the generated models of CorrFL. Moreover, it sheds light on the possibility of facilitating the convergence of local models. This property can be useful to address the straggling participants issue in the FL environment. The heterogeneity in computing resources can put off the learning process of some local models, which can be facilitated by other more powerful local agents that continuously send their updated models. Furthermore, the results with $d=1$ hint that the CorrFL approach can generate good results at the start of the training process, which means that not a lot of dispatched model weights are needed to obtain good results. The observations of the delay effect open the door for applying the CorrFL model to address different concerns in the FL environment, ranging from stragglers to intermittent connectivity issues.

\subsection{Effect of Validation Epochs}
After analyzing the effect of $d$ on the performance of $m3\_benchmark$ and $m3\_CorrFL$ models, the next step discusses the results of varying the $VE$. $VE$ determines how much of the novel environment is incorporated into the model weights of available devices. This factor involves a downstream effect on the $m3\_CorrFL$ model and the utility of the adopted approach in the studied use case.

Figures \ref{fig:10_VE_150_D} and \ref{fig:10_VE_75_D} illustrate a sample of the effect of $VE$ on the performance of the available models, $m3\_benchmark$, and $m3\_CorrFL$ model when applied to the validation set.  Additionally, the effect of $VE$ on $m3$ if it were available is denoted by $m3$ in the figure. The inclusion of $m3$ and $m3\_benchmark$ results are instrumental to provide the upper and lower bound performance. 
In particular, results in Figure \ref{fig:10_VE_150_D} are common among different combinations of $VE$ and $d$ parameters, whereas the observed phenomenon in Figure \ref{fig:10_VE_75_D} represents an outlier. Both cases are included because they provide interesting insights and trigger intriguing discussions. 

Figure \ref{fig:10_VE_150_D} depicts the variation of MAE with the $VE$ for $d = 10$. With regard to the available models $m1$ and $m2$, their respective MAE decreases slightly with the increase in the number of epochs.  Similar observation is drawn out for $m3$.  The biggest drop in MAE is attained after the first epoch, suggesting that the model weights are adjusted to the novel environment. After the first epoch, no noticeable gains in performance are acquired. Similar observations are extracted for Figure \ref{fig:10_VE_75_D}; however, a more prominent decrease is observed for the $m1$ to an MAE that resembles the one in Figure \ref{fig:10_VE_150_D}. The dynamics of the CorrFL model performance are slightly different as a result of the increase in $VE$. In Figure \ref{fig:10_VE_150_D}, the $m3\_CorrFL$ performance steadily improves and stabilizes at epoch 5 to deteriorate slightly after that epoch. This variation shows that beyond a specific epoch, the generated CorrFL weights are less reflective of the novel environment. A minimal dissimilarity in this dynamic is observed for Figure \ref{fig:10_VE_75_D}, whereby the inflection point is at epoch 2. Therefore, a sweet spot exists that balances performance gains while avoiding the significant increase in PI. As the number of $VE$ increases, the PI increase, which diminishes any performance improvement by the $m3\_CorrFL$.  The inclusion of $m3$ in Figure  \ref{fig:10_VE_150_D} shows that there is room for performance improvement for the generated CorrFL model. This improvement can be achieved by applying 
a hyper-parameter optimization procedure to the CorrFL model and by experimenting with a wider range of parameters, an aspect that was not touched upon in the current manuscript.

The CorrFL-generated models significantly outperformed the benchmark models in all combinations, except for the one depicted in Figure \ref{fig:10_VE_75_D}. Under the same assumptions of $d$ and $VE$, the CorrFL models have better performance as illustrated in Figure \ref{fig:train_test_75}. This outlier can be attributed to the possible slow convergence of the available models during the training phase. One of the models may have been stuck in a local minimum, which engendered model weights of low quality and minimal correlation with respect to the other models. Additionally, the $m3\_benchmark$ outperforms all of its counterparts, which implies that the model converged to its best performance, a condition predicated on the initialization of weights. Moreover, in this case, the underlying assumption is that each model was trained on the training data for 5 epochs ($d=5$). This case is one of the many possible delays that can be encountered in such an environment.

The provided discussion misses a very important factor in the field of model retraining during the validation phase, which can explain the underwhelming results in some cases. This phenomenon is referred to as catastrophic forgetting \cite{french1999catastrophic}, which when projected to the studied environment, assumes that the model forgot the dynamics under normal conditions. Some of these dynamics can be successfully translated to high activity level conditions; however, the training on the validation set contributed to forgetting these dynamics. This phenomenon is not studied in this manuscript, and it will be investigated in future work. 
\begingroup
\setlength{\tabcolsep}{6pt} 
\renewcommand{\arraystretch}{1.2}
\begin{table}[]
\centering
\caption{Effect of Frequency Models on MAE and PI for $VE = 5$}
\begin{tabular}{|c|c|c|c|}
\hline
\textbf{Delay}      & \textbf{Model Dispatch Frequency}  & \textbf{MAE} & \textbf{PI (Mb)} \\ \hline
\multirow{3}{*}{1} & 5                         & 167.53       & 0.09             \\ \cline{2-4} 
                    & 10                        & 225.52       & 0.11             \\ \cline{2-4} 
                    & 15                        & 155.66       & 0.03             \\ \hline
\end{tabular}

\label{tab:fm_15}
\end{table}
\endgroup

\subsection{Effect of Model Dispatch Frequency}
 
The importance of analyzing the effect of Model Dispatch Frequency ($MDF$) stems from its impact on the training data size used by the CorrFL model. As such, studying this impact sheds light on the CorrFL model's ability to train its AEs so that they neither overfit nor underfit the models. To that end, Table \ref{tab:fm_15} only includes the results for $d = 1$, given that this delay produces the least amount of training data, compared to other delay values. Therefore, analyzing the alteration of this parameter allows for a better understanding of the amount of data required to produce satisfactory results for the $m3\_CorrFL$. An additional advantage is gauging the networking resources to allocate for the realization of the adopted architecture, especially if the studied IoT devices are deployed in a harsh environment with limited access to bandwidth resources.  

Table \ref{tab:fm_15} summarizes the variation of MAE and PI with respect to changing the $MDF$. While the table only involves $d = 1$, similar observations are reported for other delays. There is no prominent performance trend with the increase of $MDF$. However, the best-performing models are obtained with $MDF = 15$ outperforming other $MDF$ values in both the MAE and PI parameters. This observation can be attributed to two main factors. First, decreasing the $MDF$ means an automatic increase in the data size, which the CorrFL can easily overfit, despite requiring a substantial amount of data to converge. Second, increasing the $MDF$ results in the collection of coarser model weights from each IoT node. This modification yields more disparate models, instead of the repetitive ones in the finer-grained scenario. Under these circumstances, the CorrFL model is fed with a more diverse dataset that favours the realization of weights that model the normal and high activity levels case environments. 

\begin{figure}
    \centering
    \includegraphics[scale=0.25]{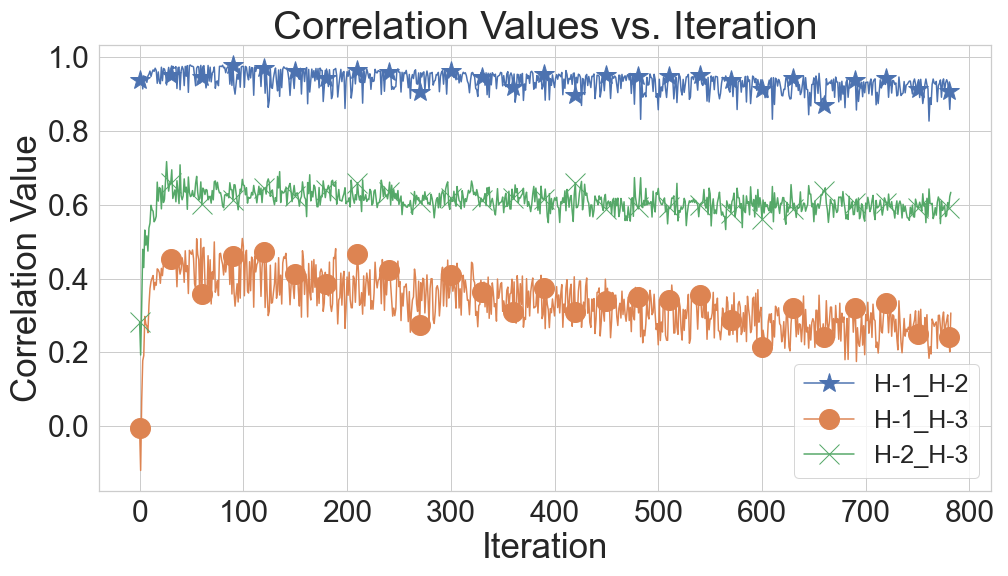}
    \caption{Correlation Analysis}
    \label{fig:correlation_analysis}
\end{figure}

\subsection{Correlation Analysis of CorrFL}
This subsection is devoted to analyzing the correlation between hidden representations $H_i$ obtained from the CorrFL model. This analysis highlights that the CorrFL approach has contributions that extend beyond the FL paradigm into NN similarity inference. Additionally, it presents deeper insights into the similar trajectory that models with shared feature space follow in their training process. Furthermore, in terms of relevance to FL, it alludes to the deduction of the data heterogeneity aspects of the models and the aspirations for model weight compression. Since the models share some of the feature space, it is expected that they share some neuron combinations. However, previous studies have shown that each NN forms its own unique set of features to realize its desired task. Figure \ref{fig:correlation_analysis} depicts the common representation $H_i$ variation for the CorrFL model across one epoch of training, encompassing 783 iterations. The notation in the figure follows this pattern $H-{i}\_H-{j}$ such that $i$ and $j$ are the indices of the absent models. 

The highest correlation is observed for the combination encompassing $W_1$ and $W_2$. This result means that the remainder of the models when these models are absent are capable of capturing common representations. This commonality shows that the data gathered by each IoT device is superfluous and the model weight compression can be applied. These trends are less magnified in the two other cases. Specifically, the common representations when $m1$ or $m3$ are absent have lower correlations. This means that their absence is providing unique information that is captured by the common representation.

\subsection{Time and Space Complexity Analysis}
\begin{figure*}
    \centering
    \begin{subfigure}[t]{0.45\textwidth}
        \centering
        \includegraphics[scale=0.3]{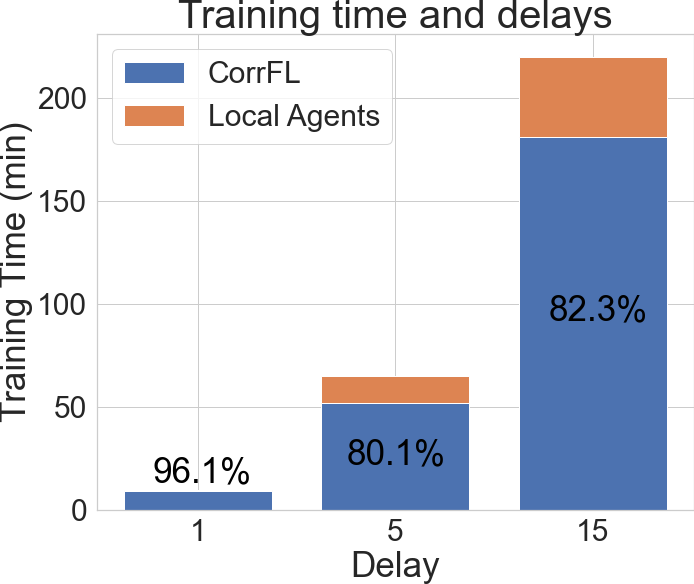}
        \caption{$MDF = 5$}
        \label{fig:training_time}
    \end{subfigure}%
    ~ 
    \begin{subfigure}[t]{0.45\textwidth}
        \centering
        \includegraphics[scale=0.25]{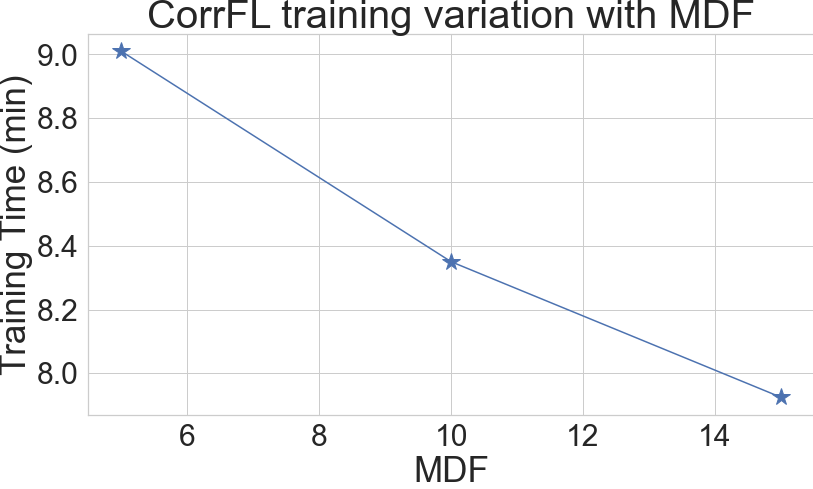}
        \caption{$d = 5$}
        \label{fig:MDF_training_time}
    \end{subfigure}
    \caption{Variation of local agents' and CorrFL model's training time}
\end{figure*}
This section analyzes the training time of local agents' models and the server model represented by the CorrFL model and the training data size per delay ($d$) and inference data size for the CorrFL model. Delay ($d$) and Model Dispatch Frequency ($MDF$) are the main contributing factors to the variation in training times. On the one hand, $d$ determines the amount of training data for the CorrFL model and the number of epochs for the local agents. On the other hand, $MDF$ controls the number of models sent at once to train the CorrFL model, which also contributes to the increase or shrinkage of its corresponding training data. As a result, the training time of local agents is analyzed in light of parameter $d$ while the training time of the CorrFL model is investigated based on $d$ and $MDF$. 

Figures \ref{fig:training_time} and \ref{fig:MDF_training_time} depict the variations in training time with respect to $MDF$ and  $d$. Figure \ref{fig:training_time} shows the effect of $d$ on the training time with a constant $MDF = 5$ and the percentage of CorrFL training time to the sum of the training time of Local Agent and CorrFL models. The local agents' training can be executed in parallel, which means that the depicted values represent the average of all the involved local agents. As expected, the training time of CorrFL and local agents increases with the increase of $d$. This positive correlation is attributed to the increase in training data for both parties. A noticeable drop in the contribution of CorrFL's training time with the increase in $d$ to 5 to increase slightly for $d = 15 $ epochs. This drop is caused by the non-linearity in the increase in training data between the CorrFL and the local agent's models. When the $d$ increases from 1 to 5, the local agents' training data significantly surged compared to a lesser increase for CorrFL models. This trend is curbed with the further increase in $d$ from 5 to 15, which means the training data proportional increase did not significantly change. Figure \ref{fig:MDF_training_time} shows the variation of CorrFL's training time for $d = 5$ epochs and $MDF$ values of 5, 10, and 15. Since the CorrFL's training data shrink with less frequent model dispatch, the training time is expected to drop with the increase in $MDF$. The inference times for local agent models and CorrFL models are negligible within 0.02 seconds for the testing phase, which highlights the applicability of the defined approach.  

As for the space complexity, the analysis covers the data size required for the training and inference of the CorrFL model. This model takes three inputs, each with 448, 336, and 448 features representing the weights between the input layer and the first hidden layer. The inference time requires a single data point for each input data, amounting to 0.009 Mbs of data. Similar to the CorrFL's training time, the training data size is dependent on $d$ and $MDF$. In that regard, the training data size is reported per epoch ($d = 1$) and $MDF = 5$. For $MDF = 5$ and $d = 1$, a single $CC$ constituting 20, 160 data points generates 504 weights for each model. In a single epoch ($d = 1$), $CC = 15$, which is equivalent to 7560 data points. Considering that each CorrFL training iteration requires three models, the input data represents a matrix of $7560 \times 448$, $7560 \times 336$, and $7560 \times 448$ respectively for models $m1$, $m2$, and $m3$. As a result, the size of training data with the defined configuration is equivalent to $74.51 Mbs$ of exchanged data. The extrapolation of data size with respect to different $d$ and $MDF$ is a straightforward exercise. Accordingly, it is imperative to calibrate the $MDF$ parameter based on the computational and communication resources available in the studied environment. 

\section{Conclusion}
The distributed IoT environment presents a challenge to the conventional centralized approaches to gathering data and applying ML techniques for automation systems. Therefore, FL is proposed as a collaborative method to address the salient issues of the centralized approach. However, the local model's heterogeneity and availability constraints that are encountered in real-world conditions hamper the realization of the envisioned FL models.  Together, these challenges and the research questions that follow are referred to as ``Oblique Federated Learning". This work devises the CorrFL approach to jointly address these practical hurdles. CorrFL is applied to a use case involving the prediction of CO\textsubscript{2} concentrations in a specific time horizon. The adopted approach is evaluated in a use case with a sudden increase in occupants' activity levels, which is directly linked to CO\textsubscript{2} predictions and the unavailability of one of the models. The results show that the model weights outputted by CorrFL outperform the benchmark models in CO\textsubscript{2} predictions. While the initial results are satisfactory, the evaluation of this approach spawned many research questions. These questions include the optimization of different parameters and addressing the possibility of catastrophic forgetting. Future work will address all of these questions.

\bibliographystyle{IEEEtran}
\bibliography{refs}
\newpage

\begin{wrapfigure}{l}{23mm} 
    \includegraphics[width=1in,height=1.25in,clip,keepaspectratio]{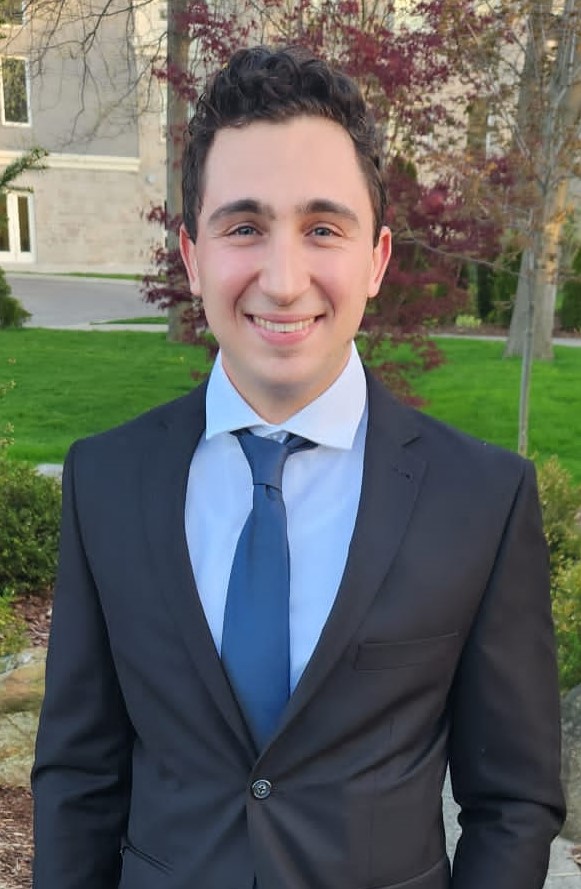}
  \end{wrapfigure}
\textbf{Ibrahim Shaer} received his B.S. degree in Computer Science from the American University of Beirut, Beirut, Lebanon in 2017 and his M.E.Sc. degree in Electrical and Computer Engineering from the University of Western Ontario, London, Canada in 2020. He is pursuing his Ph.D. in Electrical and Computer Engineering from the University of Western Ontario, London, Canada as part of the Optimized Computing and Communication Laboratory. His research interests include applications of Machine Learning in industrial buildings, such as the optimization of Heating, Ventilation, and Air Conditioning systems and anomaly detection and Machine Learning interpretation in high-dimensional spaces. He is an active IEEE volunteer and a member of the IEEE Computer Society.

\begin{wrapfigure}{l}{23mm} 
    \includegraphics[width=1in,height=1.25in,clip,keepaspectratio]{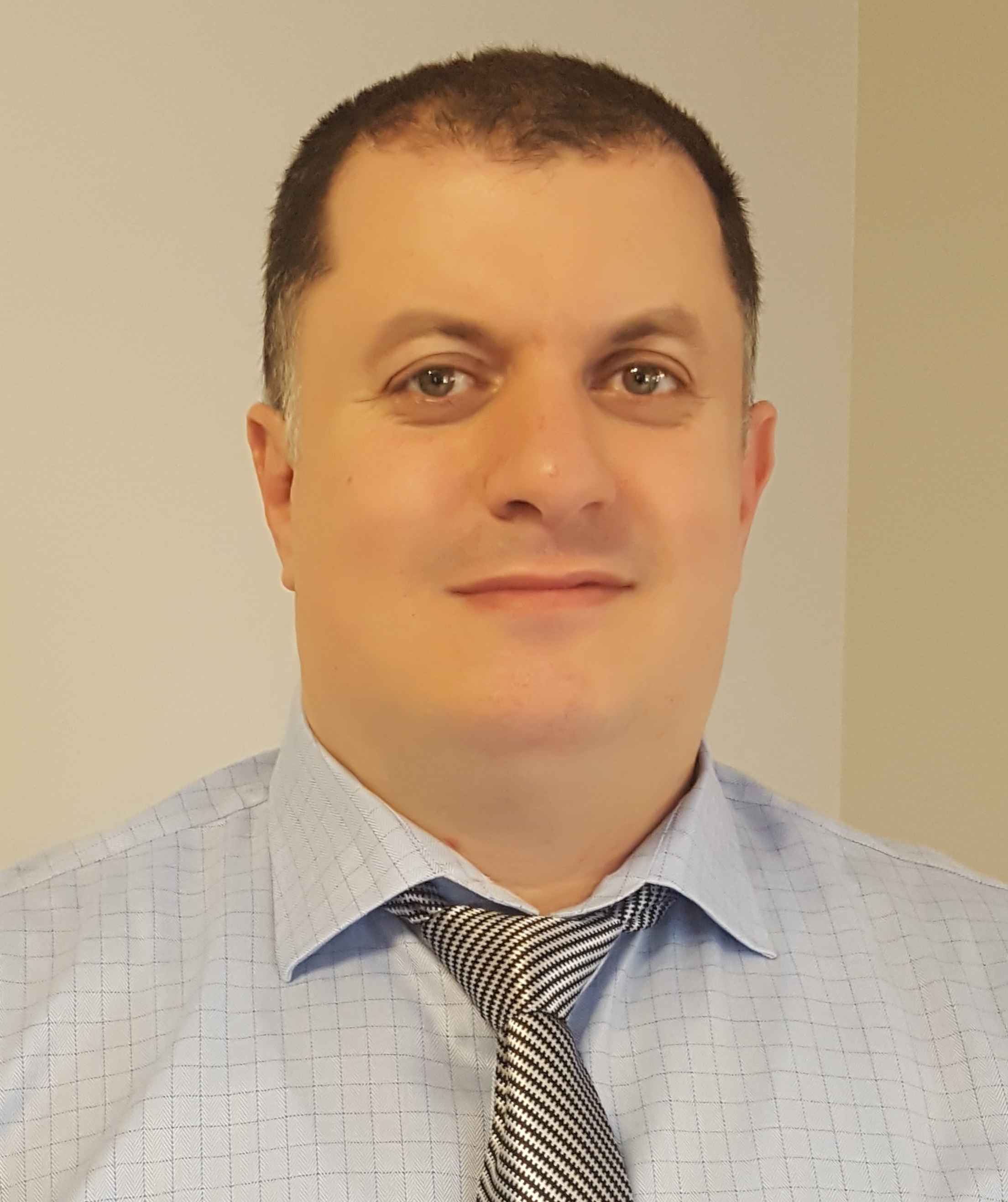}
  \end{wrapfigure}
\textbf{Abdallah Shami} received his B.E. degree in Electrical and Computer engineering from Lebanese University, Beirut, Lebanon, in 1997, and his Ph.D. degree in Electrical and Computer engineering from the Graduate School and University Center, City University of New York, New York, NY, USA, in 2003. He is the Acting Associate Dean of Research and a Professor at the ECE department at Western University, Ontario, Canada. Dr. Shami’s research interests are in the area of future networks, the Internet of Things, and smart systems. He is currently an Associate Editor for IEEE Transactions on Mobile Computing, IEEE Internet of Things Journal, and IEEE Communications Tutorials and Survey. He was the elected Chair of the IEEE Communications Society Technical Committee on Communications Software and IEEE London Ontario Section Chair.

\end{document}